\documentclass{article}

\usepackage[preprint]{neurips_2025}

\usepackage[utf8]{inputenc}
\usepackage[T1]{fontenc}
\usepackage{hyperref}
\usepackage{url}
\usepackage{booktabs}
\usepackage{multirow}
\usepackage{amsfonts}
\usepackage{graphicx}
\usepackage{nicefrac}
\usepackage{microtype}
\usepackage{algorithm}
\usepackage{amsmath}
\usepackage{enumitem}
\usepackage[x11names,table]{xcolor} 
\usepackage{gensymb}
\usepackage{algpseudocode}

\title{Ascending the Infinite Ladder: Benchmarking Spatial Deformation Reasoning in Vision-Language Models}

\author{Jiahuan Zhang\textsuperscript{1,2}\thanks{~~Equal contribution}\hspace{4pt}\thanks{~~Project leader}\hspace{4pt}, 
Shunwen Bai\textsuperscript{1,3}\footnotemark[1], 
Tianheng Wang\textsuperscript{1}\footnotemark[1], 
Kaiwen Guo\textsuperscript{1,4}\footnotemark[1],\\
\textbf{Kai Han\textsuperscript{5}},  
\textbf{Guozheng Rao\textsuperscript{2}},
\textbf{Kaicheng Yu\textsuperscript{1}\thanks{~~Corresponding author}}\\
    \textsuperscript{1} Autolab, Westlake University~~
    \textsuperscript{2} Tianjin University~~
    \textsuperscript{3} Zhejiang University\\
    \textsuperscript{4} Capital Normal University~~
    \textsuperscript{5} University of Hong Kong\\
    \texttt{\{zhangjiahuan78, kyu\}@westlake.edu.cn}
}

\begin{document}

\maketitle

\begin{figure}[h]
    \centering
      \includegraphics[width=1\linewidth]{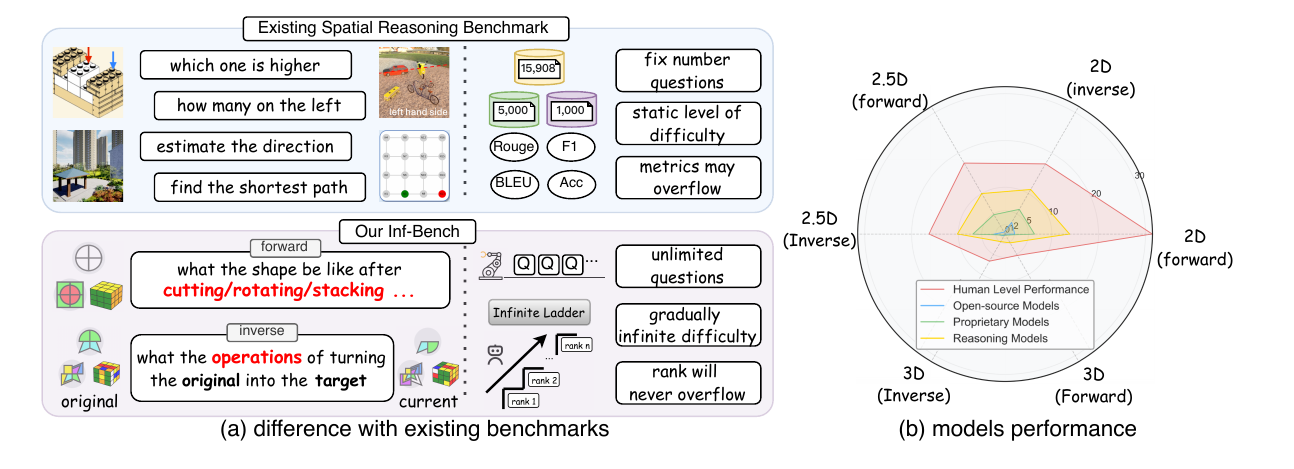}
    \caption{
Existing spatial reasoning benchmarks~\cite{tang2025lego,zhan2025open3dvqa,wang2025pulsecheck457,tang_sparkle_2025} for Vision-Language Models (VLMs) focus on tasks like `which is higher' or `find the shortest path,' with a static level of difficulty, which
may easily get outdated with the fast evolving of VLMs. Our Inf-Bench introduces tasks requiring forward and reverse \textbf{spatial deformation reasoning} (e.g., shape changes after cutting/rotating/stacking) across 2D-3D. Our \textbf{ladder competition} offers evolving tasks, ensuring continuous capability exploration without performance saturation. The results show that even the best reasoning models still lag \textbf{significantly behind humans} in spatial deformation reasoning.
    }
    \label{fig: teaser} 
\end{figure}

\begin{abstract}
Humans naturally possess the spatial reasoning ability to form and manipulate images and structures of objects in space. There is an increasing effort to endow Vision-Language Models (VLMs) with similar spatial reasoning capabilities. However, it remains unclear whether these models truly understand and manipulate spatial objects or not. To address this question, we propose a new evaluation framework aimed at assessing the performance of VLMs in spatial deformation reasoning tasks. Specifically, we construct a benchmark for spatial deformation reasoning from 2D to 3D. Leveraging our data engine, we can generate unlimited evaluation problem pairs with infinite steps, without any data leakage. We explore whether the model can effectively perform spatial deformation reasoning from two directions: forward reasoning (given the operations, find the final state) and reverse reasoning (given the final state, determine the operations). We adopt a ladder competition format, using the number of deformation steps as the level classification criterion, with the goal of exploring the boundaries of the model's deformation reasoning capabilities. Interestingly, the benchmarking results reveal that almost no model demonstrates plausible spatial deformation reasoning abilities. Furthermore, even after applying targeted training and mainstream reasoning enhancement methods, the models are still unable to perform well on 3D spatial deformation reasoning. 

\end{abstract}
\section{Introduction}
Imagine shaping a lump of clay into a desired object through sequential manipulations, while considering its shape and topological relationship. Each action requires predicting how the clay will deform and planning the process to achieve a smooth, consistent outcome. This involves spatial deformation~\cite{gain2008survey}, a common task in daily life, such as modeling with clay or crafting sculptures. While current Vision-Language Models (VLMs) excel in language understanding and image recognition~\cite{hou2024vision, hong2024cogvlm2, chen2024evlm, xia2024vision, bordes2024introduction, ghosh2024exploring, li2025benchmark, pandey2025benchmarking, ho2025review}, the question remains: \textit{can they effectively handle complex spatial deformation tasks?}

Existing benchmarks explore aspects of three-dimensional spatial reasoning and visual-language reasoning in dynamic environments, they predominantly focus on static or dynamic scenes and spatial relationships between objects~\cite{mayer2025ivispar, tang2025lego, wang2025pulsecheck457, xu2025visulogic, yang2024thinking, zhan2025open3dvqa, tang_sparkle_2025}, as shown in Figure~\ref{fig: teaser}. However, no benchmark executes in-depth evaluations of spatial deformation, specifically how models handle the dynamic transformation of objects' shapes in space.

To fill this gap and find the limits of models' deformation reasoning, we introduce Inf-Bench, integrating the Shapez game~\cite{springer2020shapez} and the Rubik's Cube to test VLMs' deformation reasoning ability. These games are selected for their simple rules and emphasis on object manipulation, free from prior knowledge or pattern reliance. We examine two reasoning types: forward reasoning (transforming objects step-by-step from the initial to the target state) and inverse reasoning (inferring the steps needed to reach the target from the end state). We also create a fully automated data engine to produce unlimited and diverse evaluation tasks, ensuring no data leakage.
 
We propose an ``Infinite Ladder Competition,'' where the number of required deformation steps defines task difficulty. All models start at one-step tasks and advance to more complex ones as they succeed. Unlike traditional benchmarks (shown in Figure~\ref{fig: teaser}) with fixed tasks and metrics, which become obsolete once models achieve perfect scores \cite{mayer2025ivispar,xu2025visulogic, oliveira_sliding_2025}, our system allows unlimited scalability. Model performance is reflected by the highest level completed, enabling continuous, dynamic evaluation of reasoning capabilities within a flexible and extensible framework.

We evaluate several mainstream VLMs and find that human performance outperforms all models across tasks, particularly in 3D reasoning, where most models, even the powerful model OpenAI o3 \cite{openai2024o3o4mini}, fail to complete even basic reasoning tasks. Further analysis reveal that the models primarily rely on two strategies: ``direct pairing'' and ``step-by-step reasoning execution''. The former involves making intuitive choices based on patterns and regularities, while the latter first encodes the figure and then analyzes deformations in the encoding before mapping it back to the figure. However, the results indicate that both strategies have significant performance bottlenecks, highlighting the models' difficulties in managing high-dimensional coupling relationships. 

Additionally, we fine-tune the models using datasets of varying scales and find that in lower-dimensional tasks, the models are able to learn forward iterative transformations and extend generalizable reasoning. However, in 3D forward tasks, the reasoning depth remains limited, representing a fundamental bottleneck. At the same time, we explore several mainstream VLM enhancement methods, including Chain-of-Thought (COT) \cite{wei2022chain} and ReAct \cite{yao2023react}, among others. However, we find their effects are minimal, and in some cases, they even hinder performance. Thus, we can conclude that current VLMs still lack stable and generalizable spatial deformation reasoning abilities.  

The main contributions of this paper can be summarized as follows:
    (i) We introduce the first effective benchmark for comprehensively exploring the spatial deformation reasoning ability of VLMs.
    (ii) We propose an infinitely scalable benchmarking paradigm based on the ladder competition system.
    (iii) We provide a thoroughly evaluation and identifying the VLMs' shortcomings in spatial deformation reasoning, providing a clear direction for future model development.
\section{Spatial Deformation Reasoning}
In this study, \textbf{spatial deformation reasoning} refers to the model's ability to understand, predict, and execute complex deformations of an object's shape. We focus on implementing this reasoning in VLMs, especially when there is no prior knowledge, and the model must learn and perform deformations through observation or manipulation. Unlike \textbf{visual spatial intelligence} in the VSI-Bench~\cite{yang2024thinking}, which emphasizes localization, relational understanding, and spatial perception, our spatial deformation centers on dynamic shape transformations, particularly in multi-step processes that change an object’s state. We categorize the core capabilities of spatial deformation as follows:

\textbf{Spatial Recognition.} The ability to comprehend the initial shape of an object and accurately identify the areas that require deformation.

\textbf{Abstraction of Operational Law Principles.} The ability to grasp the principles behind deformation operations and abstract them into deformation laws, crucial for accurate reasoning.

\textbf{Stable Reasoning Execution.} The ability to gradually execute the inferred reasoning rules, maintain stability throughout the process, and derive the final state.

\section{Inf-Bench}

\subsection{Benchmark Overview}

\begin{figure}[h]
    \centering
      \includegraphics[width=0.9\linewidth]{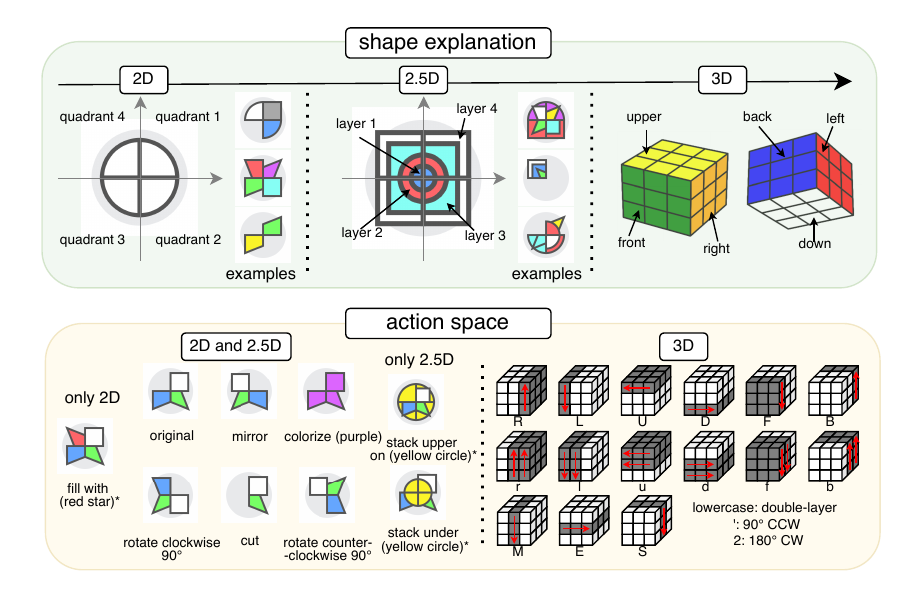}
    \caption{
    \textbf{Shape Explanation.} 2D shapes are presented on a flat plane, while 2.5D introduces additional (up to four) layers of dimension. 3D shapes, based on a Rubik's Cube, introduce depth and spatial orientation. \textbf{Action Space.} Actions like rotation and coloring are common in 2D and 2.5D, with 2D including ``filling'' and 2.5D adding ``stacking.'' The 3D section focuses on Rubik's Cube rotations and movements in three dimensions.
    }
    \label{fig: shapez_and_cube} 
\end{figure}

We introduce the Inf-Bench dataset, designed to evaluate the performance of VLMs in spatial deformation reasoning tasks. Inf-Bench covers deformation tasks from 2D to 3D, including the following three spatial levels:

\textbf{2D Tasks:} This task, inspired by single-plane deformation in Shapez, evaluates the model's ability to recognize objects and perform shape transformations in a two-dimensional plane. Shapes are single-layer figures divided into four quadrants, each containing one of four predefined patterns or an ``empty'' state (see Figure~\ref{fig: shapez_and_cube}). The task includes six deformation operations, involving cutting, rotating, or coloring to transform the overall shape. 

\textbf{2.5D Tasks:} Extending the 2D task, this task adds a vertical stacking dimension, creating a multi-layer deformation challenge. Each planar figure is expanded into a stacked structure with up to four layers  (see Figure~\ref{fig: shapez_and_cube}). The model must perform intra-layer transformations and reason about alignment and composition across layers. Seven deformation operations are defined to assess multi-layer deformation and spatial reasoning abilities. 

\textbf{3D Tasks:} Based on the Rubik's Cube, this task tests the ability to manipulate and reason spatially in three dimensions (see Figure~\ref{fig: shapez_and_cube}). The Cube has 27 units, forming 54 visible faces that can change position and orientation through rotations. The task allows all 54 basic rotations to evaluate the model's 3D spatial reasoning, action planning, and hierarchical reasoning. All the complete introductions of shapes and operations of these three dimensions' tasks are shown in the Appendix~\ref{app: more details of shapes}.

\begin{figure}[h]
    \centering
      \includegraphics[width=\linewidth]{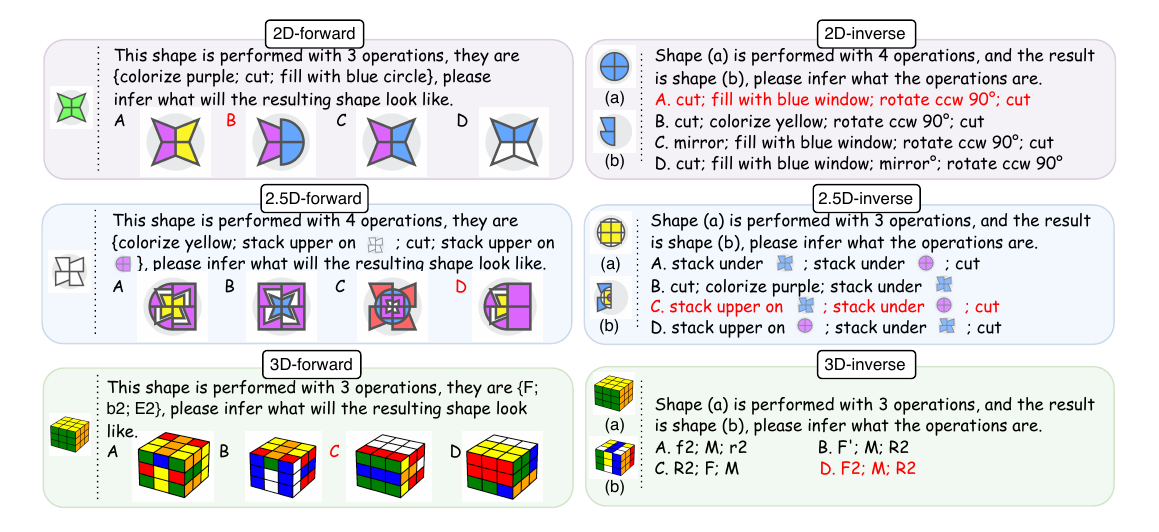}
    \caption{
    \textbf{Inf-Bench Task Examples.} Tasks are presented from 2D to 3D, with forward tasks (left) and reverse tasks (right). Note: All problems have been simplified to enhance clarity and conciseness.
    }
    \label{fig: foward_and_inverse} 
\end{figure}

Each spatial level includes two types of tasks: \textbf{Forward Reasoning} and \textbf{Inverse Reasoning}. In \textbf{Forward Reasoning Tasks}, the model transforms an object from its initial state to the target state through deformation steps. These tasks focus on how the model understands and executes the sequence and effects of spatial operations. In \textbf{Inverse Reasoning Tasks}, given the target state, the model deduces the steps to achieve the transformation, assessing the model's ability to reverse-engineer the deformation path. Task examples are shown in the Figure~\ref{fig: foward_and_inverse} and all the prompts are presented in the Appendix~\ref{app: General Evaluation Setup}. 

\subsection{Dataset Construction}

\begin{figure}[h]
    \centering
      \includegraphics[width=1\linewidth]{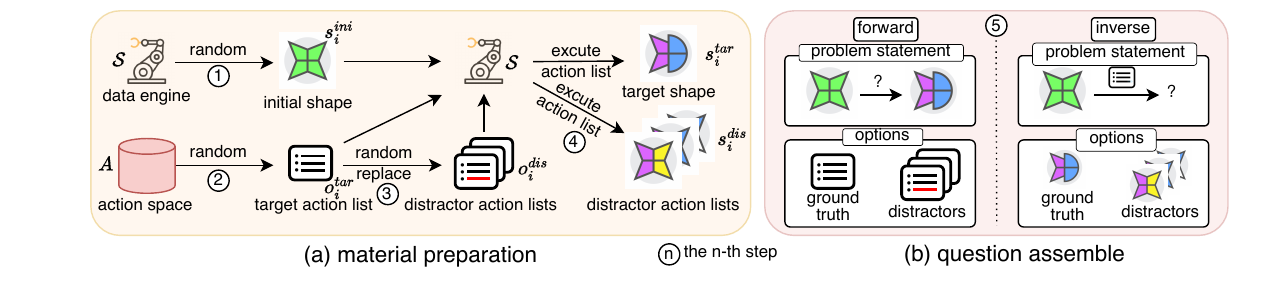}
    \caption{
    \textbf{Data Generation Pipeline.} The process consists of five steps: First, a shape is generated, and then a target deformation operation list is created. Next, an interference list is generated by modifying the target list. Target and interference shapes are then produced, and finally, materials are assembled into both forward and reverse tasks.
    }
    \label{fig: pipeline} 
\end{figure}

To ensure data quality and accuracy, we designed automated \textbf{shape generation engine}: \( \mathcal{S} \), which performs two main functions: 
1) Randomly generating an initial shape, and
2) Applying a sequence of deformations (action list) to generate the corresponding target shape. The shape generation engine is rule-based, with generation and operation execution implemented as fixed Python functions, detailed in the Appendix~\ref{app: data_engine}. The deformation execution is deterministic, ensuring that the same initial shape and action list always produce a unique and consistent final shape. This guarantees reproducibility, eliminates manual annotation errors, and enhances the dataset's reliability for evaluation.


The data generation pipeline consists of two main processes: \textbf{material preparation} and \textbf{question assembly}, which are divided into five steps: 

\textbf{Step 1}: The shape generation engine \( \mathcal{S} \) randomly selects an initial shape \( s_i^\text{ini} \), given by \( s_i^\text{ini} = \mathcal{S}(\text{Init}) \). 

\textbf{Step 2}: \( n \) actions are randomly chosen from the action space \( A \) to create the action list \( o_i^\text{tar} = \text{RandList}(n, A) \).

\textbf{Step 3}: Distractor action lists are generated by replacing selected operations in \( o_i^\text{tar} \) with new ones from \( A \), forming \( k \) distractor lists \( o_{i,j}^\text{dis} = \text{RandReplace}(o_i^\text{tar}, r, A) \), i.e., \( o_i^\text{dis} = \{ o_{i,j}^\text{dis} \}_{j=1}^{k} \). 

\textbf{Step 4}: The action lists \( o_i^\text{tar} \) and \( o_{i,j}^\text{dis} \) are applied to the initial shape \( s_i^\text{ini} \) to generate the target shape \( s_i^\text{tar} = \mathcal{S}(s_i^\text{ini}, o_i^\text{tar}) \) and distractor shapes \( s_{i,j}^\text{dis} = \mathcal{S}(s_i^\text{ini}, o_{i,j}^\text{dis}) \), i.e., \( s_i^\text{dis} = \{ s_{i,j}^\text{dis} \}_{j=1}^{k} \). 

\textbf{Step 5}: Multiple-choice questions are generated as follows. The forward question \( q_i^\text{for} \) consists of an option set \( op_i^\text{for} = \{ s_i^\text{tar}, s_i^\text{dis} \} \), and its composition is given by \( q_i^\text{for} = \left( \{ s_i^\text{ini}, o_i^\text{tar}, op_i^\text{for} \}, \, \text{gt}_i = s_i^\text{tar} \right) \). The inverse question \( q_i^\text{inv} \) has the option set \( op_i^\text{inv} = \{ o_i^\text{tar}, o_i^\text{dis} \} \), with the composition \( q_i^\text{inv} = \left( \{ s_i^\text{ini}, s_i^\text{tar}, op_i^\text{inv} \}, \, \text{gt}_i = o_i^\text{tar} \right) \), where \( \text{gt}_i \) represents the ground truth for each question.

\section{Evaluation on Inf-Bench}
\subsection{Ladder Competition Framework}
To explore the boundaries of spatial deformation reasoning abilities and systematically evaluate models' performance, we employ the ladder competition framework. The goal is progressively increasing the difficulty, providing a comprehensive test of the model's adaptability in deformation tasks. In the ladder competition, models begin at the lowest difficulty level, denoted as \( R = 1 \) with only one step deformation. Each level consists of five questions. If a model successfully answers at least three of these questions, it advances to a higher difficulty level; otherwise, it is downgraded to the same level. If it fails at the same level twice, the competition ends:
\[
R = \begin{cases} 
R - 1 \text{ and } f_R = f_R + 1 & \text{if } c < 3 \\
R + 1 & \text{if } c \geq 3 \\
R & \text{if } f_R = 2 \text{ or } R = 0
\end{cases}
\]
where \( f_R \) represents failure times of a particular \( R \), and \( c \) denotes the number of correctly answered questions at this level. This mechanism thoroughly tests the model's reasoning abilities under increasing cognitive load. The rules for advancing, downgrading, and the number of allowed failures at each level ensure the rigor of the evaluation and the stability of the model's reasoning capabilities. \textbf{The final value of \( R \), representing reasoning depth, is used as the Inf-Bench metric.}
\subsection{Evaluation Setting}
We thoroughly evaluate 18 image-supported VLMs, representing a diverse range of model families, parameter scales, training strategies, and advanced reasoning capabilities. We also recruit 100 human volunteers to conduct the testing. The proprietary models under consideration include GPT-4.1-2025-04-14 \cite{achiam2023gpt}, Claude 3.7 Sonnet \cite{anthropic2024claude}, Gemini-2.5-Flash-04-17 \cite{team2023gemini}, and Doubao 1.5 Vision Pro. For open-source models, we evaluate those from the Qwen series \cite{bai2023qwen,yang2024qwen2}, Llama 4 Scout, Llama 4 Maverick \cite{meta2025llama}, and Intern VLM \cite{chen2024expanding}. Additionally, the reasoning models incorporated into our analysis comprise O3 , O4 Mini \cite{openai2024o3o4mini}, Gemini-2.5-pro-03-25 \cite{team2023gemini}, Grok 3 Reasoning \cite{xai2025grok3}, Claude 3.7 Sonnet Thinking \cite{anthropic2024claude}, and Doubao 1.5 Thinking Pro \cite{doubao2025doubao15pro}.
All evaluations are conducted in a zero-shot setting. To ensure robustness and impartiality, each model is evaluated on the same task set across all ranks, with 10 independent runs per model for statistical validity. Greedy decoding was used for all models to enhance reproducibility. The results are shown in Table~\ref{tab:main_result}.

\begin{table}[h!]
\caption{\textbf{Evaluation on Inf-Bench.} The data shows the average reasoning depth \( R \) achieved by the models after 10 ladder competitions. \colorbox[HTML]{C0C0C0}{Light gray} indicates the best performance by open-source models, \colorbox[HTML]{D2F8DF}{light green}  by proprietary models, and \colorbox[HTML]{DAE8FC}{light blue} by reasoning models. Note that, for simplicity, the version numbers of the models (e.g., 04-13) are not included in the table.}
\label{tab:main_result}
\centering
\footnotesize
\setlength{\tabcolsep}{2pt}
\begin{tabular}{c|cc|cc|cc|c|cc|cc|cc}
\toprule
 & \multicolumn{2}{c}{2D} & \multicolumn{2}{c}{2.5D} & \multicolumn{2}{c}{3D} &  & \multicolumn{2}{c}{2D} & \multicolumn{2}{c}{2.5D} & \multicolumn{2}{c}{3D} \\
 Models & For & Inv & For & Inv & For & Inv & Models & For & Inv & For & Inv & For & Inv \\
\midrule
Human & 31.5 & 17.3 & 17.5 & 16.3 & 6.7 & 5.3 &  &  &  &  &  &  &  \\
\midrule
\multicolumn{14}{c}{Open-source Models} \\
Qwen2.5-VL-7B & 0.9 & 0.1 & 0.1 & 0.5 & 0.3 & 0.0 & InternVL3-14B & 0.8 & 0.1 & 0.1 & 0.4 & 0.4 & 0.0 \\
Qwen2.5-VL-32B & 2.2 & 4.4 & 0.2 & 2.9 & 0.1 & 0.0 & InternVL3-38B & 2.0 & 3.2 & 0.1 & 3.5 & 0.1 & 0.0 \\
Qwen2.5-VL-72B & 3.0 & 5.1 & 0.1 & \cellcolor[HTML]{C0C0C0}4.4 & 0.6 & 0.0 & InternVL3-78B & 3.5 & 4.8 & 0.1 & 3.8 & 0.7 & 0.0 \\
llama-4-scout & 0.4 & 0.6 & 0.6 & 0.6 & 0.8 & 0.0 & llama-4-maverick & \cellcolor[HTML]{C0C0C0}3.6 & \cellcolor[HTML]{C0C0C0}5.2 & \cellcolor[HTML]{C0C0C0}3.0 & 2.9 & \cellcolor[HTML]{C0C0C0}1.0 & 0.0 \\
\midrule
\multicolumn{14}{c}{Proprietary Models} \\
GPT-4.1-nano & 0.0 & 1.0 & 0.0 & 1.0 & 0.1 & 0.0 & claude-3-7-sonnet & 8.9 & 6.0 & 2.3 & 8.6 & 0.4 & 0.0 \\
GPT-4.1-mini & 4.6 & 1.9 & 3.9 & 4.4 & 0.0 & 0.0 & Gemini-2.5-flash-preview & \cellcolor[HTML]{D2F8DF}11.5 & \cellcolor[HTML]{D2F8DF}17.5 & \cellcolor[HTML]{D2F8DF}11.5 & \cellcolor[HTML]{D2F8DF}16.0 & 0.2 & \cellcolor[HTML]{D2F8DF}1.6 \\
GPT-4.1 & 8.6 & 7.6 & 6.6 & 6.4 & \cellcolor[HTML]{D2F8DF}1.2 & 0.0 & doubao-1-5-vision-pro & 4.5 & 2.9 & 4.5 & 4.7 & 0.0 & 0.1 \\
\midrule
\multicolumn{14}{c}{Reasoning Models} \\
O3 & \cellcolor[HTML]{DAE8FC}27.0 & \cellcolor[HTML]{DAE8FC}15.3 & \cellcolor[HTML]{DAE8FC}13.6 & 10.4 & \cellcolor[HTML]{DAE8FC}4.2 & 3.9 & grok-3-reasoning & 10.9 & 11.7 & 8.5 & 10.9 & 1.7 & 2.5 \\
O4-mini & 16.7 & 10.0 & 8.4 & 6.8 & 2.4 & 2.5 & claude-3-7-sonnet-thinking & 7.7 & 8.4 & 5.8 & 5.2 & 1.1 & 0.4 \\
Gemini-2.5-pro & 14.5 & 14.9 & 12.4 & \cellcolor[HTML]{DAE8FC}15.0 & 1.5 & \cellcolor[HTML]{DAE8FC}4.1 & doubao-1.5-thinking-pro & 5.9 & 5.6 & 11.5 & 12.7 & 0.7 & 0.0\\
\bottomrule
\end{tabular}
\end{table}

\subsection{Main Results}
\textbf{Human-Level Performance.} As expected, the final rank scores of human evaluators all exceed those of the best-performing models, whether in 2D, 2.5D, or 3D tasks. They generally perform better in forward reasoning tasks than reverse reasoning tasks.

\textbf{Model Performance Overview.} It is evident that models generally show a competitive performance in 2D and 2.5D tasks. It is also found that the performance of the reasoning model and proprietary models is significantly better than that of the open-source models. However, in tasks that require actual 3D spatial transformation reasoning, both open-source and proprietary models struggle to perform even a single step of reasoning. This highlights a significant limitation in their spatial deformation reasoning capabilities. Only powerful reasoning models, such as O3, O4-mini, and Gemini-2.5-pro-03-25, demonstrate some initial 3D deformation abilities. However, their performance still lags significantly behind human capabilities, indicating substantial room for improvement.


\section{Discussion}
\subsection{How Do VLMs Perform on Spatial Deformation Reasoning?}

\begin{figure}[h]
    \centering
      \includegraphics[width=0.95\linewidth]{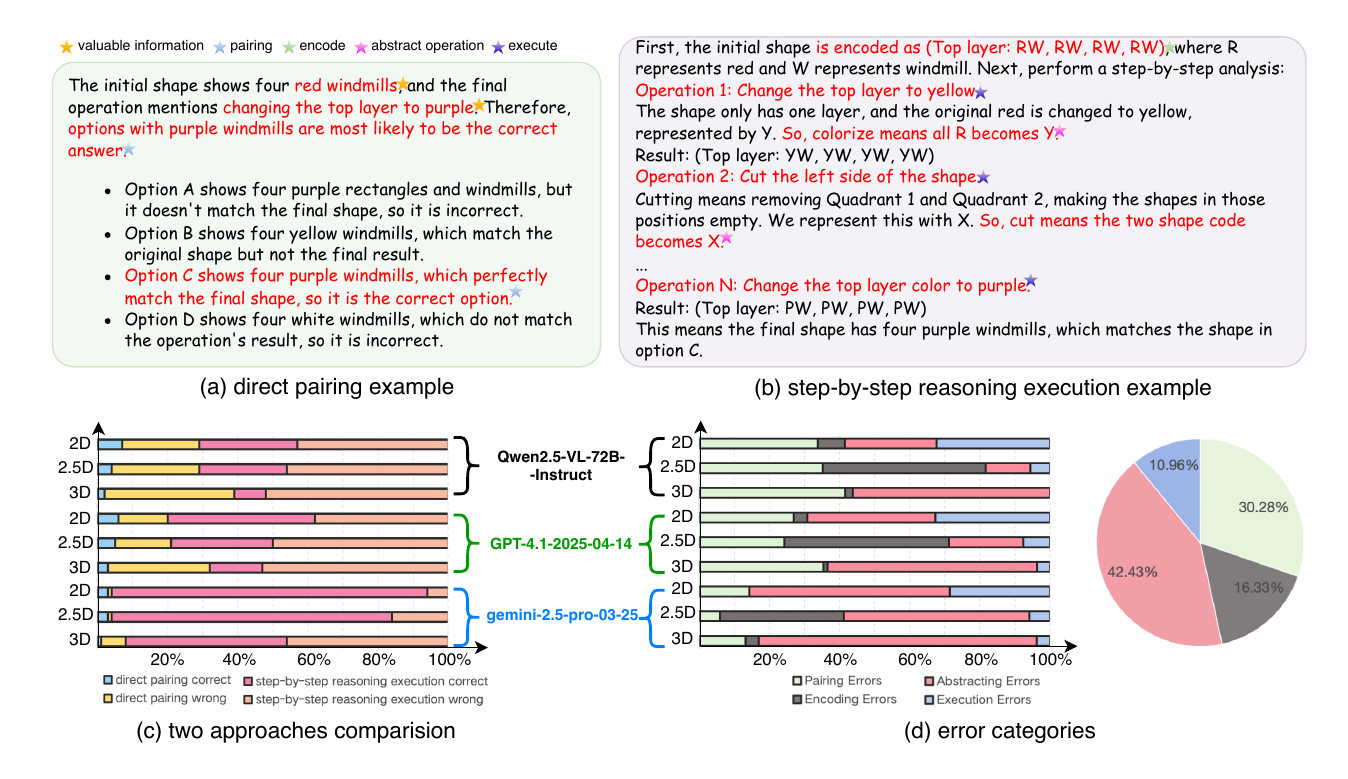}
    \caption{
    \textbf{Case Study.} (a) illustrates the direct pairing method, where valuable information is extracted from the prompt and used to pair options directly. (b) shows the example of step-by-step reasoning execution, where the model first encodes the statement, abstracts actions into operations, and then executes them sequentially. (c) compares the proportions of the two methods, highlighting that more complex tasks favor the step-by-step reasoning approach, which also tends to have higher execution accuracy. (d) displays the distribution of main error types made by models.
    }
    \label{fig: case} 
\end{figure}

To better understand model performance, we analyze the internal reasoning processes to reveal the mechanisms. We select representative models from three categories: Qwen2.5-VL-72B-Instruct, GPT-4.1, and Gemini-2.5-pro-03-25, and collect their reasoning processes. Each collect 100 responses per task, totaling 900 responses. Through analysis, we find that the model primarily utilizes two approaches to response. One is \textbf{direct pairing} (see Figure~\ref{fig: case} (a)), where the model extracts valuable information directly from the operations and pairs them with the options. The second is \textbf{step-by-step reasoning} (see Figure~\ref{fig: case} (b)), where the model encodes the input shapes and abstracts various operations into a set of standardized encoding rules. The model then executes the operations based on these rules to derive the final answer. From Figure~\ref{fig: case} (c), it is evident that stronger models derive more conclusions through step-by-step reasoning. As difficulty increases, the proportion of direct pairings also rises, reflecting the model's tendency to infer from valuable information in complex tasks.

We analyze 502 incorrect responses from 900 and categorize the errors into four types, as shown in Figure~\ref{fig: case} (d). Full error analysis and examples are in Appendix~\ref{app: error_ana}.

\textbf{Pairing Errors} occur when models incorrectly pair based on wrong information from the problem statement, accounting for 30.28\% of errors. 

\textbf{Encoding Errors} arise when models confuse shape encoding, especially with multi-layered shapes, leading to incorrect answers, representing 16.33\%. 

\textbf{Abstracting Errors} happen when models misinterpret operations or their effects, accounting for 42.43\%, particularly in cube tasks where coupled shape states complicate abstraction. Lastly, 

\textbf{Execution Errors} occur during multi-step reasoning, with one step incorrectly executed, making up 10.96\% of errors. These are more common in non-reasoning models.

\subsection{How Do VLMs Perform on Spatial Deformation Reasoning with Only Encoded Input?}
\label{main:5-3}

\begin{figure}[h!]
    \centering
      \includegraphics[width=1\linewidth]{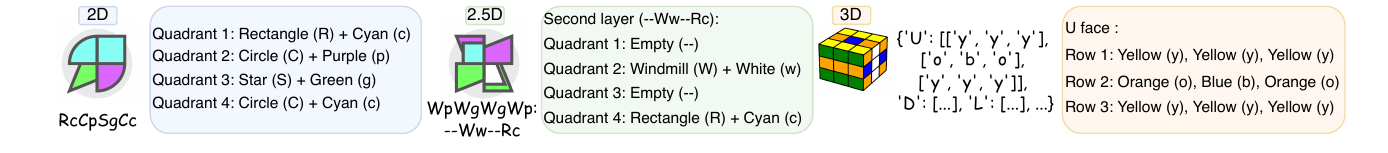}
    \caption{
    \textbf{Encoded Shape Example.}
    }
    \label{fig: encode_example} 
\end{figure}

As noted earlier, models primarily rely on shape encoding for spatial deformation reasoning. To minimize perceptual encoding errors, we pre-encode all images uniformly, allowing models to process pure text input for reasoning. The encoding methods are detailed in the Appendix~\ref{app: encode_method}, examples in Figure~\ref{fig: encode_example} and results in Table~\ref{tab:result_encoded}. 

\begin{table}[h!]
\caption{\textbf{Evaluation on Inf-Bench (only Encoded Input).} The data shows the average reasoning depth \( R \) achieved by the models after 10 ladder competitions. \colorbox[HTML]{C0C0C0}{Light gray} indicates the best performance by open-source models, \colorbox[HTML]{D2F8DF}{light green}  by proprietary models, and \colorbox[HTML]{DAE8FC}{light blue} by reasoning models. Note that, for simplicity, the version numbers of the models (e.g., 04-13) are not included.}
\label{tab:result_encoded}
\centering
\footnotesize
\setlength{\tabcolsep}{1.5pt}
\begin{tabular}{c|cc|cc|cc|c|cc|cc|cc}
\toprule
 & \multicolumn{2}{c}{2D} & \multicolumn{2}{c}{2.5D} & \multicolumn{2}{c}{3D} &  & \multicolumn{2}{c}{2D} & \multicolumn{2}{c}{2.5D} & \multicolumn{2}{c}{3D} \\
Models & For & Inv & For & Inv & For & Inv & Models & For & Inv & For & Inv & For & Inv \\
\midrule
Human & 7.2 & 5.7 & 6.4 & 4.1 & 1.9 & 1.3 &  &  &  &  &  &  &  \\
\midrule
\multicolumn{14}{c}{Open-source Models} \\
Qwen2.5-VL-7B & 2.1 & 0.3 & 2.1 & 2.5 & 0.0 & 0.0 & InternVL3-14B & 2.4 & 0.4 & 1.8 & 2.1 & 0.1 & 0.0 \\
Qwen2.5-VL-32B & 23.7 & 3.1 & 10.9 & 7.4 & 0.0 & 0.0 & InternVL3-38B & 12.7 & 3.7 & 8.9 & \cellcolor[HTML]{C0C0C0}8.4 & 0.3 & 0.0 \\
Qwen2.5-VL-72B & 25.0 & \cellcolor[HTML]{C0C0C0}6.2 & 11.8 & \cellcolor[HTML]{C0C0C0}8.4 & 0.1 & 0.0 & InternVL3-78B & \cellcolor[HTML]{C0C0C0}29.7 & 5.4 & \cellcolor[HTML]{C0C0C0}12.5 & 7.2 & 0.3 & 0.0 \\
llama-4-scout & 0.1 & 0.0 & 1.0 & 0.9 & 0.0 & 0.0 & llama-4-maverick & 4.1 & 3.0 & 3.2 & 2.2 & \cellcolor[HTML]{C0C0C0}1.4 & \cellcolor[HTML]{C0C0C0}0.2 \\
\midrule
\multicolumn{14}{c}{Proprietary Models} \\
GPT-4.1-nano & 6.3 & 1.6 & 2.5 & 1.8 & 0.3 & 0.0 & claude-3-7-sonnet & 53.1 & 11.0 & 15.8 & 18.6 & 1.5 & 0.0 \\
GPT-4.1-mini & 52.3 & 10.8 & 15.2 & 11.8 & 0.5 & 0.0 & Gemini-2.5-flash-preview & \cellcolor[HTML]{D2F8DF}137.5 & \cellcolor[HTML]{D2F8DF}65.9 & \cellcolor[HTML]{D2F8DF}16.7 & \cellcolor[HTML]{D2F8DF}25.6 & \cellcolor[HTML]{D2F8DF}5.6 & \cellcolor[HTML]{D2F8DF}5.6 \\
GPT-4.1 & 63.3 & 11.1 & 16.9 & 12.2 & 1.6 & 0.0 & doubao-1-5-vision-pro & 16.1 & 7.5 & 10.8 & 12.0 & 0.1 & 0.1 \\
\midrule
\multicolumn{14}{c}{Reasoning Models} \\
O3 & \cellcolor[HTML]{DAE8FC}671.4 & 61.6 & \cellcolor[HTML]{DAE8FC}61.1 & \cellcolor[HTML]{DAE8FC}17.6 & 4.4 & 4.2 & grok-3-reasoning & 393.2 & 23.1 & 25.4 & 12.9 & 4.1 & 2.8 \\
O4-mini & 433.2 & 17.5 & 19.7 & 15.2 & 3.8 & 2.9 & claude-3-7-sonnet-thinking & 32.2 & 9.6 & 6.7 & 16.9 & 0.7 & 0.0 \\
Gemini-2.5-pro & 562.6 & \cellcolor[HTML]{DAE8FC}121.3 & 46.1 & 16.9 & \cellcolor[HTML]{DAE8FC}5.1 & \cellcolor[HTML]{DAE8FC}5.9 & doubao-1.5-thinking-pro & 92.8 & 21.1 & 21.1 & 22.3 & 2.0 & 0.0 \\
\bottomrule
\end{tabular}
\end{table}

The key findings are that models exhibit significant improvement in 2D and 2.5D tasks, however, they still face challenges in 3D tasks, specifically:

\textbf{Performance in 2D and 2.5D tasks improves significantly,} suggesting that models effectively abstract operational patterns for tasks involving operations like cutting and rotating, which typically lack complex coupled states. Reasoning models show even greater improvements, with O3 achieving forward reasoning depth even over 671.4, and Gemini-2.5-pro-03-25 reaching an inverse reasoning depth of 121.3, demonstrating stable performance.

\textbf{However, in 3D tasks, most models still struggle with operations on 3D shapes}, where each operation involves interactions between coupled shape states. For instance, Rubik's Cube operations require considering interactions across multiple faces. These multidimensional dependencies increase reasoning complexity. Only high-performance models like Gemini-2.5-pro-03-25 offer partial solutions, with reasoning depth \( R \) of 6, indicating room for improvement.

\textbf{Interestingly, human performance on text-based tasks is lower than visual input}, often lagging behind the models, highlighting the difference in thinking processes between humans and models.

\subsection{Can  Supervised Fine-tuning Effectively Enhance Spatial Deformation Reasoning Ability?}


\begin{table}[h!]
\centering
\caption{\textbf{Evaluation on Inf-Bench After SFT.} The data shows the average \( R \) of 10 ladder challenges, after SFT. $S_{\text{max}}$ represents the highest difficulty level in the fine-tuning data. ``For'' refers to forward reasoning training, while ``Inv'' refers to inverse reasoning.
}
\label{tab:sft_result}
\footnotesize
\setlength{\tabcolsep}{2.5pt}
\begin{tabular}{c|cc|cc|cc|c|cc|cc|cc}
\toprule
 & \multicolumn{2}{c}{2D} & \multicolumn{2}{c}{2.5D} & \multicolumn{2}{c}{3D} &  & \multicolumn{2}{c}{2D} & \multicolumn{2}{c}{2.5D} & \multicolumn{2}{c}{3D} \\
 & For & Inv & For & Inv & For & Inv &  & For & Inv & For & Inv & For & Inv \\
\midrule
\multicolumn{7}{c}{Qwen2.5-VL-7B-Instruct} & \multicolumn{7}{c}{Qwen2.5-VL-72B-Instruct} \\
Vanilia & 0.9 & 0.1 & 0.1 & 0.5 & 0.3 & 0.0 & Vanilia & 3.0 & 5.1 & 0.1 & 4.4 & 0.6 & 0.0 \\
\midrule
For (\( S_{\text{max}} \) = 1) & 3.6 & 4.0 & 5.2 & 11.6 & 1.2 & 0.0 & For (\( S_{\text{max}} \) = 1) & 4.2 & 5.7 & 7.5 & 12.4 & 1.3 & 0.0 \\
Inv (\( S_{\text{max}} \) = 1) & 4.0 & 4.4 & 3.5 & 10.2 & 1.0 & 1.0 & Inv (\( S_{\text{max}} \) = 1) & 5.0 & 7.4 & 5.4 & 13.7 & 1.2 & 1.2 \\
\midrule
For (\( S_{\text{max}} \) = 5) & 11.9 & 5.7 & 13.3 & 12.6 & 6.4 & 3.8 & For (\( S_{\text{max}} \) = 5) & 33.2 & 13.4 & 12.7 & 23.7 & 6.1 & 5.0 \\
Inv (\( S_{\text{max}} \) = 5) & 6.7 & 74.4 & 6.9 & 43.2 & 0.8 & 65.2 & Inv (\( S_{\text{max}} \) = 5) & 12.0 & 373.6 & 7.9 & 264.4 & 2.1 & 79.0 \\
\midrule
For (\( S_{\text{max}} \) = 10) & 39.8 & 11.2 & 14.1 & 21.3 & 6.6 & 4.7 & For (\( S_{\text{max}} \) = 10) & 47.6 & 14.4 & 15.4 & 27.1 & 7.1 & 5.6 \\
Inv (\( S_{\text{max}} \) = 10) & 8.2 & 166.5 & 9.4 & 149.6 & 1.9 & 134.9 & Inv (\( S_{\text{max}} \) = 10) & 12.3 & 415.6 & 10.8 & 349.8 & 2.6 & 267.2 \\
\bottomrule
\end{tabular}
\end{table}

The above discussion shows that existing models have significant room for improvement in spatial deformation reasoning. This leads to the key question: Can the classical training paradigm enable models to tackle this issue? To investigate, we conduct a supervised fine-tuning (SFT) experiment with two open-source models, Qwen2.5-VL-7B-Instruct and Qwen2.5-VL-72B-Instruct, which perform poorly in initial tests and differ in parameter scale. We use the \( S_{\text{max}} \) to classify the datasets, where \( S_{\text{max}} \) represents the most difficult data covered in each dataset. Each group contains 20,000 samples, with equal data across steps from 1 to \( S_{\text{max}} \). Training details are in the Appendix~\ref{app: train_detail}

The key findings are that SFT notably enhances reasoning depth and generalization, but in 3D tasks, the forward task's performance remains limited, revealing the shortcomings of traditional training methods for high-dimensional deformation reasoning, specifically:

\textbf{Inverse task achieves deeper inference, forward task limited.} In the forward task, reasoning depth improves with increasing \( S_{\text{max}} \) but remains limited in 3D tasks. In contrast, the inverse task is more robust, reducing error accumulation by gradually verifying the target state, which enhances solution depth, especially as \( S_{\text{max}} \) increases.

\textbf{SFT enhances 2D/2.5D depth, 3D forward task limited.} SFT significantly boosts reasoning depth in 2D and 2.5D tasks, enabling multi-step reasoning beyond training examples, suggesting the learning of a reusable deformation operator. However, in 3D tasks, even with high-difficulty data (e.g., \( S_{\text{max}} = 10 \)), the forward task struggles to exceed five steps, highlighting the difficulty of capturing 3D deformation sequences and error accumulation with only supervised examples.

\textbf{Forward task shows stronger generalization than inverse task.} Models trained with the forward task show better generalization in the backward task than those trained with the inverse task, indicating that stepwise reasoning in the forward task aids solving the inverse task, while the lack of forward reasoning limits inverse task models' generalization.

\begin{table}[h!]
\centering
\caption{\textbf{Evaluation on Inf-Bench After Reasoning Enhancement.} The data represents the difference in the mean highest rank achieved by the models after 10 ladder challenges using the reasoning enhancement method, compared to the baseline performance (i.e., vanilla results).}
\label{tab:enhance_result}
\footnotesize
\setlength{\tabcolsep}{3pt}
\begin{tabular}{c|cc|cc|cc|c|cc|cc|cc}
\toprule
 & \multicolumn{2}{c}{2D} & \multicolumn{2}{c}{2.5D} & \multicolumn{2}{c}{3D} & \multicolumn{1}{l}{} & \multicolumn{2}{c}{2D} & \multicolumn{2}{c}{2.5D} & \multicolumn{2}{c}{3D} \\
 & For & Inv & For & Inv & For & Inv & \multicolumn{1}{l}{} & For & Inv & For & Inv & For & Inv \\
\midrule
\multicolumn{7}{c}{GPT-4.1} & \multicolumn{7}{c}{llama-4-scout} \\
Vanilia & 8.6 & 7.6 & 6.6 & 6.4 & 1.2 & 0.0 & Vanilia & 0.4 & 0.6 & 0.6 & 0.6 & 0.8 & 0.0 \\
\midrule
+COT & -0.1 & -0.4 & -0.7 & +0.1 & -0.2 & 0.0 & +COT & +0.3 & -0.0 & -0.1 & -0.1 & -0.3 & 0.0 \\
+few-shot & +0.6 & +0.3 & +0.4 & +0.1 & +0.0 & 0.0 & +few-shot & +0.5 & +0.1 & -0.0 & -0.0 & -0.1 & 0.0 \\
+self-reflection & -5.1 & -1.9 & -1.1 & +0.2 & -0.8 & 0.0 & +self-reflection & +0.1 & -0.1 & +0.0 & +0.1 & +0.2 & 0.0 \\
+ReAct & -3.6 & -2.7 & -6.3 & -0.5 & -0.8 & 0.0 & +ReAct & +0.0 & -0.1 & -0.2 & -0.0 & -0.1 & 0.0 \\
+Tools & -0.4 & -0.3 & -1.6 & +1.9 & -1.1 & +0.1 & +Tools & +0.1 & +0.1 & -0.0 & -0.0 & -0.1 & 0.0\\
\bottomrule
\end{tabular}
\end{table}

\subsection{Can Reasoning Enhancement Methods Improve Spatial Deformation Reasoning?}
\label{main:enhance_method}
We explore several mainstream reasoning enhancement methods to assess whether they could effectively improve spatial deformation reasoning, including Chain-of-Thought (COT)~\cite{wei2022chain}, Few-shot learning, Self-reflection, Tool Invocation, and Reasoning and Action (ReAct) \cite{yao2023react}. Implementation details can be found in the Appendix~\ref{app: enhancement_detail}. The results (see Table~\ref{tab:enhance_result}) indicate that these methods led to slight improvements in specific tasks and dimensions, but overall, \textbf{they do not significantly enhance the ability to perform spatial deformation reasoning.} This is especially evident in 3D tasks and backward tasks, where the effects are more limited.

\section{Related Work}
\textbf{VLMs Spatial Reasoning Benchmark. }Recent studies highlight spatial reasoning as key for embodied intelligence. Despite progress in multimodal learning, VLMs still struggle with complex spatial tasks. Interactive benchmarks like iVISPAR \cite{mayer2025ivispar} assess planning and spatial awareness, showing gaps in VLM performance compared to humans. SPGym \cite{oliveira_sliding_2025} exposes limitations in generalization across visual inputs, while LEGO-Puzzles \cite{tang2025lego} shows VLMs answering only half of spatial reasoning questions, compared to 90\% human accuracy. ThreeDWorld \cite{gan_threedworld_2021} helps evaluate spatial reasoning in interactive settings. VisuLogic \cite{xu2025visulogic} reveals VLMs' weak performance on spatial relations, while LLM-SRBench \cite{shojaee_llm-srbench_2025} challenges models beyond memorization. Sparkle \cite{tang_sparkle_2025} improves performance through targeted supervision, and VSI-Bench \cite{yang2024thinking} shows that generating cognitive maps enhances spatial abilities in VLMs. Our work fills the gap in 2D-to-3D spatial deformation reasoning with structured mechanics and a ladder competition format that identifies limitations.

\textbf{VLMs Reasoning Improvement.} The reasoning capabilities in LLMs~\cite{openai2024o1} stem from extensive training datasets and improved methodologies. VLMs have built on this by demonstrating visual reasoning through learning the relationships between visual inputs and text ~\cite{cheng2024videollama, li2024llava, liu2024improved, liu2023visual, wang2024internvideo2}. Motivated by LLM advancements, several studies have aimed to enhance VLM reasoning. Approaches include CoT~\cite{wei2022chain, zhang2024improve, mitra2024compositional}, ToT~\cite{yao2023tree}, GoT~\cite{besta2024graph}, and Monte Carlo methods~\cite{trinh2024solving, wan2024alphazero, wu2025boosting}, along with SFT datasets to improve reasoning performance~\cite{ye2025limo, muennighoff2025s1, yao2024mulberry, thawakar2025llamav}. Despite progress, spatial understanding, such as geometry and relationships, remains a challenge for VLMs~\cite{liu2024mmbench, ramakrishnan2025does, qiu2025can, yang2024thinking}. This has sparked research to improve spatial abilities through pre-training~\cite{xia2025geox}, 3D data construction~\cite{cai2024spatialbot, chen2024spatialvlm, huang2024rekep, ray2024sat, tang2025lego}, reasoning frameworks~\cite{cheng2024spatialrgpt}, multi-modal alignment~\cite{ray2024sat}, and Chain-of-Thought methods~\cite{liu2025spatialcot, wu2024mind, li2025imagine, tang2025lego}. Despite these efforts, spatial imagination and deformation remain underexplored. 
\section{Conclusion}
We introduce Inf-Bench to evaluate VLMs' performance in spatial deformation reasoning. The results highlight significant limitations in handling complex 3D tasks, with VLMs struggling in multi-step reasoning, despite using reasoning enhancement methods. Human performance consistently surpasses all models, emphasizing the gap between current VLMs and human spatial imagination. Our findings suggest that improvements are needed in error reduction, multi-step abstract reasoning, and 3D reasoning capabilities. This study contributes to the future development of robust spatial reasoning frameworks and identifies key directions for improving VLM architectures. Despite our comprehensive assessment of VLMs' spatial deformation reasoning capabilities, this study exhibits inherent limitations, which are discussed in detail in the Appendix~\ref{app: limitations}.

\bibliographystyle{plain}
\bibliography{references}

\begin{thebibliography}{10}

\bibitem{achiam2023gpt}
Josh Achiam, Steven Adler, Sandhini Agarwal, Lama Ahmad, Ilge Akkaya,
  Florencia~Leoni Aleman, Diogo Almeida, Janko Altenschmidt, Sam Altman,
  Shyamal Anadkat, et~al.
\newblock Gpt-4 technical report.
\newblock {\em arXiv preprint arXiv:2303.08774}, 2023.

\bibitem{anthropic2024claude}
Anthropic.
\newblock The claude 3 model family: Opus, sonnet, haiku.
\newblock Anthropic Blog, 5 2024.
\newblock https://www.anthropic.com/claude.

\bibitem{bai2023qwen}
Jinze Bai, Shuai Bai, Yunfei Chu, Zeyu Cui, Kai Dang, Xiaodong Deng, Yang Fan,
  Wenbin Ge, Yu~Han, Fei Huang, et~al.
\newblock Qwen technical report.
\newblock {\em arXiv preprint arXiv:2309.16609}, 2023.

\bibitem{besta2024graph}
Maciej Besta, Nils Blach, Ales Kubicek, Robert Gerstenberger, Michal
  Podstawski, Lukas Gianinazzi, Joanna Gajda, Tomasz Lehmann, Hubert
  Niewiadomski, Piotr Nyczyk, et~al.
\newblock Graph of thoughts: Solving elaborate problems with large language
  models.
\newblock In {\em Proceedings of the AAAI Conference on Artificial
  Intelligence}, volume~38, pages 17682--17690, 2024.

\bibitem{bordes2024introduction}
Florian Bordes, Richard~Yuanzhe Pang, Anurag Ajay, Alexander~C Li, Adrien
  Bardes, Suzanne Petryk, Oscar Ma{\~n}as, Zhiqiu Lin, Anas Mahmoud, Bargav
  Jayaraman, et~al.
\newblock An introduction to vision-language modeling.
\newblock {\em arXiv preprint arXiv:2405.17247}, 2024.

\bibitem{cai2024spatialbot}
Wenxiao Cai, Iaroslav Ponomarenko, Jianhao Yuan, Xiaoqi Li, Wankou Yang, Hao
  Dong, and Bo~Zhao.
\newblock Spatialbot: Precise spatial understanding with vision language
  models.
\newblock {\em arXiv preprint arXiv:2406.13642}, 2024.

\bibitem{chen2024spatialvlm}
Boyuan Chen, Zhuo Xu, Sean Kirmani, Brain Ichter, Dorsa Sadigh, Leonidas
  Guibas, and Fei Xia.
\newblock Spatialvlm: Endowing vision-language models with spatial reasoning
  capabilities.
\newblock In {\em Proceedings of the IEEE/CVF Conference on Computer Vision and
  Pattern Recognition}, pages 14455--14465, 2024.

\bibitem{chen2024evlm}
Kaibing Chen, Dong Shen, Hanwen Zhong, Huasong Zhong, Kui Xia, Di~Xu, Wei Yuan,
  Yifei Hu, Bin Wen, Tianke Zhang, et~al.
\newblock Evlm: An efficient vision-language model for visual understanding.
\newblock {\em arXiv preprint arXiv:2407.14177}, 2024.

\bibitem{chen2024expanding}
Zhe Chen, Weiyun Wang, Yue Cao, Yangzhou Liu, Zhangwei Gao, Erfei Cui, Jinguo
  Zhu, Shenglong Ye, Hao Tian, Zhaoyang Liu, et~al.
\newblock Expanding performance boundaries of open-source multimodal models
  with model, data, and test-time scaling.
\newblock {\em arXiv preprint arXiv:2412.05271}, 2024.

\bibitem{cheng2024spatialrgpt}
An-Chieh Cheng, Hongxu Yin, Yang Fu, Qiushan Guo, Ruihan Yang, Jan Kautz,
  Xiaolong Wang, and Sifei Liu.
\newblock Spatialrgpt: Grounded spatial reasoning in vision language models.
\newblock {\em arXiv preprint arXiv:2406.01584}, 2024.

\bibitem{cheng2024videollama}
Zesen Cheng, Sicong Leng, Hang Zhang, Yifei Xin, Xin Li, Guanzheng Chen,
  Yongxin Zhu, Wenqi Zhang, Ziyang Luo, Deli Zhao, et~al.
\newblock Videollama 2: Advancing spatial-temporal modeling and audio
  understanding in video-llms.
\newblock {\em arXiv preprint arXiv:2406.07476}, 2024.

\bibitem{gain2008survey}
James Gain and Dominique Bechmann.
\newblock A survey of spatial deformation from a user-centered perspective.
\newblock {\em ACM Transactions on Graphics (TOG)}, 27(4):1--21, 2008.

\bibitem{gan_threedworld_2021}
Chuang Gan, Jeremy Schwartz, Seth Alter, Damian Mrowca, Martin Schrimpf, James
  Traer, Julian~De Freitas, Jonas Kubilius, Abhishek Bhandwaldar, Nick Haber,
  Megumi Sano, Kuno Kim, Elias Wang, Michael Lingelbach, Aidan Curtis, Kevin
  Feigelis, Daniel~M. Bear, Dan Gutfreund, David Cox, Antonio Torralba,
  James~J. DiCarlo, Joshua~B. Tenenbaum, Josh~H. McDermott, and Daniel L.~K.
  Yamins.
\newblock {ThreeDWorld}: a platform for interactive multi-modal physical
  simulation, December 2021.
\newblock arXiv:2007.04954 [cs].

\bibitem{ghosh2024exploring}
Akash Ghosh, Arkadeep Acharya, Sriparna Saha, Vinija Jain, and Aman Chadha.
\newblock Exploring the frontier of vision-language models: A survey of current
  methodologies and future directions.
\newblock {\em arXiv preprint arXiv:2404.07214}, 2024.

\bibitem{ho2025review}
Huu-Tuong Ho, Luong~Vuong Nguyen, Minh-Tien Pham, Quang-Huy Pham, Quang-Duong
  Tran, Duong Nguyen~Minh Huy, and Tri-Hai Nguyen.
\newblock A review on vision-language-based approaches: Challenges and
  applications.
\newblock {\em Computers, Materials \& Continua}, 82(2), 2025.

\bibitem{hong2024cogvlm2}
Wenyi Hong, Weihan Wang, Ming Ding, Wenmeng Yu, Qingsong Lv, Yan Wang, Yean
  Cheng, Shiyu Huang, Junhui Ji, Zhao Xue, et~al.
\newblock Cogvlm2: Visual language models for image and video understanding.
\newblock {\em arXiv preprint arXiv:2408.16500}, 2024.

\bibitem{hou2024vision}
Yifan Hou, Buse Giledereli, Yilei Tu, and Mrinmaya Sachan.
\newblock Do vision-language models really understand visual language?
\newblock {\em arXiv preprint arXiv:2410.00193}, 2024.

\bibitem{huang2024rekep}
Wenlong Huang, Chen Wang, Yunzhu Li, Ruohan Zhang, and Li~Fei-Fei.
\newblock Rekep: Spatio-temporal reasoning of relational keypoint constraints
  for robotic manipulation.
\newblock In {\em 8th Annual Conference on Robot Learning}, 2024.

\bibitem{li2024llava}
Bo~Li, Yuanhan Zhang, Dong Guo, Renrui Zhang, Feng Li, Hao Zhang, Kaichen
  Zhang, Peiyuan Zhang, Yanwei Li, Ziwei Liu, et~al.
\newblock Llava-onevision: Easy visual task transfer.
\newblock {\em arXiv preprint arXiv:2408.03326}, 2024.

\bibitem{li2025imagine}
Chengzu Li, Wenshan Wu, Huanyu Zhang, Yan Xia, Shaoguang Mao, Li~Dong, Ivan
  Vuli{\'c}, and Furu Wei.
\newblock Imagine while reasoning in space: Multimodal
  visualization-of-thought.
\newblock {\em arXiv preprint arXiv:2501.07542}, 2025.

\bibitem{li2025benchmark}
Zongxia Li, Xiyang Wu, Hongyang Du, Huy Nghiem, and Guangyao Shi.
\newblock Benchmark evaluations, applications, and challenges of large vision
  language models: A survey.
\newblock {\em arXiv preprint arXiv:2501.02189}, 1, 2025.

\bibitem{liu2024improved}
Haotian Liu, Chunyuan Li, Yuheng Li, and Yong~Jae Lee.
\newblock Improved baselines with visual instruction tuning.
\newblock In {\em Proceedings of the IEEE/CVF Conference on Computer Vision and
  Pattern Recognition}, pages 26296--26306, 2024.

\bibitem{liu2023visual}
Haotian Liu, Chunyuan Li, Qingyang Wu, and Yong~Jae Lee.
\newblock Visual instruction tuning.
\newblock {\em Advances in neural information processing systems},
  36:34892--34916, 2023.

\bibitem{liu2024mmbench}
Yuan Liu, Haodong Duan, Yuanhan Zhang, Bo~Li, Songyang Zhang, Yike Yuan, Wangbo
  Zhao, Jiaqi Wang, Conghui He, Ziwei Liu, Kai Chen, and Dahua Lin.
\newblock {MMB}ench: Is your multi-modal model an all-around player?, 2024.

\bibitem{liu2025spatialcot}
Yuecheng Liu, Dafeng Chi, Shiguang Wu, Zhanguang Zhang, Yaochen Hu, Lingfeng
  Zhang, Yingxue Zhang, Shuang Wu, Tongtong Cao, Guowei Huang, et~al.
\newblock Spatialcot: Advancing spatial reasoning through coordinate alignment
  and chain-of-thought for embodied task planning.
\newblock {\em arXiv preprint arXiv:2501.10074}, 2025.

\bibitem{mayer2025ivispar}
Julius Mayer, Mohamad Ballout, Serwan Jassim, Farbod~Nosrat Nezami, and Elia
  Bruni.
\newblock ivispar--an interactive visual-spatial reasoning benchmark for vlms.
\newblock {\em arXiv preprint arXiv:2502.03214}, 2025.

\bibitem{meta2025llama}
AI~Meta.
\newblock The llama 4 herd: The beginning of a new era of natively multimodal
  ai innovation, april 2025, 2025.

\bibitem{mitra2024compositional}
Chancharik Mitra, Brandon Huang, Trevor Darrell, and Roei Herzig.
\newblock Compositional chain-of-thought prompting for large multimodal models.
\newblock In {\em Proceedings of the IEEE/CVF Conference on Computer Vision and
  Pattern Recognition}, pages 14420--14431, 2024.

\bibitem{muennighoff2025s1}
Niklas Muennighoff, Zitong Yang, Weijia Shi, Xiang~Lisa Li, Li~Fei-Fei,
  Hannaneh Hajishirzi, Luke Zettlemoyer, Percy Liang, Emmanuel Cand{\`e}s, and
  Tatsunori Hashimoto.
\newblock s1: Simple test-time scaling.
\newblock {\em arXiv preprint arXiv:2501.19393}, 2025.

\bibitem{oliveira_sliding_2025}
Bryan L. M.~de Oliveira, Murilo L.~da Luz, Bruno Brandão, Luana G.~B. Martins,
  Telma W. de~L. Soares, and Luckeciano~C. Melo.
\newblock Sliding puzzles gym: a scalable benchmark for state representation in
  visual reinforcement learning, February 2025.
\newblock arXiv:2410.14038 [cs].

\bibitem{openai2024o1}
OpenAI.
\newblock Openai o1 system card.
\newblock OpenAI Website, 12 2024.
\newblock https://cdn.openai.com/o1-system-card-20241205.pdf, Accessed:
  2025-05-14.

\bibitem{openai2024o3o4mini}
OpenAI.
\newblock Openai o3 and o4-mini system card.
\newblock OpenAI Website, 2024.
\newblock https://openai.com/index/o3-o4-mini-system-card/, Accessed:
  2025-05-14.

\bibitem{pandey2025benchmarking}
Saugat Pandey and Alvitta Ottley.
\newblock Benchmarking visual language models on standardized visualization
  literacy tests.
\newblock {\em arXiv preprint arXiv:2503.16632}, 2025.

\bibitem{qiu2025can}
Zeju Qiu, Weiyang Liu, Haiwen Feng, Zhen Liu, Tim~Z. Xiao, Katherine~M.
  Collins, Joshua~B. Tenenbaum, Adrian Weller, Michael~J. Black, and Bernhard
  Sch{\"o}lkopf.
\newblock Can large language models understand symbolic graphics programs?
\newblock In {\em The Thirteenth International Conference on Learning
  Representations}, 2025.

\bibitem{ramakrishnan2025does}
Santhosh~Kumar Ramakrishnan, Erik Wijmans, Philipp Kraehenbuehl, and Vladlen
  Koltun.
\newblock Does spatial cognition emerge in frontier models?
\newblock In {\em The Thirteenth International Conference on Learning
  Representations}, 2025.

\bibitem{ray2024sat}
Arijit Ray, Jiafei Duan, Reuben Tan, Dina Bashkirova, Rose Hendrix, Kiana
  Ehsani, Aniruddha Kembhavi, Bryan~A Plummer, Ranjay Krishna, Kuo-Hao Zeng,
  et~al.
\newblock Sat: Spatial aptitude training for multimodal language models.
\newblock {\em arXiv preprint arXiv:2412.07755}, 2024.

\bibitem{shojaee_llm-srbench_2025}
Parshin Shojaee, Ngoc-Hieu Nguyen, Kazem Meidani, Amir~Barati Farimani, Khoa~D.
  Doan, and Chandan~K. Reddy.
\newblock {LLM}-{SRBench}: a new benchmark for scientific equation discovery
  with large language models, April 2025.
\newblock arXiv:2504.10415 [cs].

\bibitem{springer2020shapez}
Tobias Springer.
\newblock Shapez.
\newblock \url{https://store.steampowered.com/app/1318690/shapez/}, 2020.
\newblock PC, Released June 7, 2020.

\bibitem{tang2025lego}
Kexian Tang, Junyao Gao, Yanhong Zeng, Haodong Duan, Yanan Sun, Zhening Xing,
  Wenran Liu, Kaifeng Lyu, and Kai Chen.
\newblock Lego-puzzles: How good are mllms at multi-step spatial reasoning?
\newblock {\em arXiv preprint arXiv:2503.19990}, 2025.

\bibitem{tang_sparkle_2025}
Yihong Tang, Ao~Qu, Zhaokai Wang, Dingyi Zhuang, Zhaofeng Wu, Wei Ma, Shenhao
  Wang, Yunhan Zheng, Zhan Zhao, and Jinhua Zhao.
\newblock Sparkle: mastering basic spatial capabilities in vision language
  models elicits generalization to spatial reasoning, March 2025.
\newblock arXiv:2410.16162 [cs].

\bibitem{doubao2025doubao15pro}
Doubao Team.
\newblock Doubao 1.5 pro.
\newblock Doubao Website, 2025.
\newblock URL: \url{https://team.doubao.com/en/special/doubao_1_5_pro},
  Accessed: 2025-05-14.

\bibitem{team2023gemini}
Gemini Team, Rohan Anil, Sebastian Borgeaud, Jean-Baptiste Alayrac, Jiahui Yu,
  Radu Soricut, Johan Schalkwyk, Andrew~M Dai, Anja Hauth, Katie Millican,
  et~al.
\newblock Gemini: a family of highly capable multimodal models.
\newblock {\em arXiv preprint arXiv:2312.11805}, 2023.

\bibitem{thawakar2025llamav}
Omkar Thawakar, Dinura Dissanayake, Ketan More, Ritesh Thawkar, Ahmed Heakl,
  Noor Ahsan, Yuhao Li, Mohammed Zumri, Jean Lahoud, Rao~Muhammad Anwer, et~al.
\newblock Llamav-o1: Rethinking step-by-step visual reasoning in llms.
\newblock {\em arXiv preprint arXiv:2501.06186}, 2025.

\bibitem{trinh2024solving}
Trieu~H Trinh, Yuhuai Wu, Quoc~V Le, He~He, and Thang Luong.
\newblock Solving olympiad geometry without human demonstrations.
\newblock {\em Nature}, 625(7995):476--482, 2024.

\bibitem{wan2024alphazero}
Ziyu Wan, Xidong Feng, Muning Wen, Stephen~Marcus McAleer, Ying Wen, Weinan
  Zhang, and Jun Wang.
\newblock Alphazero-like tree-search can guide large language model decoding
  and training.
\newblock In {\em Forty-first International Conference on Machine Learning},
  2024.

\bibitem{wang2025pulsecheck457}
Xingrui Wang, Wufei Ma, Tiezheng Zhang, Celso~M de~Melo, Jieneng Chen, and Alan
  Yuille.
\newblock Pulsecheck457: A diagnostic benchmark for 6d spatial reasoning of
  large multimodal models.
\newblock {\em arXiv e-prints}, pages arXiv--2502, 2025.

\bibitem{wang2024internvideo2}
Yi~Wang, Kunchang Li, Xinhao Li, Jiashuo Yu, Yinan He, Guo Chen, Baoqi Pei,
  Rongkun Zheng, Zun Wang, Yansong Shi, et~al.
\newblock Internvideo2: Scaling foundation models for multimodal video
  understanding.
\newblock In {\em European Conference on Computer Vision}, pages 396--416.
  Springer, 2024.

\bibitem{wei2022chain}
Jason Wei, Xuezhi Wang, Dale Schuurmans, Maarten Bosma, Fei Xia, Ed~Chi, Quoc~V
  Le, Denny Zhou, et~al.
\newblock Chain-of-thought prompting elicits reasoning in large language
  models.
\newblock {\em Advances in neural information processing systems},
  35:24824--24837, 2022.

\bibitem{wu2025boosting}
Jinyang Wu, Mingkuan Feng, Shuai Zhang, Ruihan Jin, Feihu Che, Zengqi Wen, and
  Jianhua Tao.
\newblock Boosting multimodal reasoning with mcts-automated structured
  thinking.
\newblock {\em arXiv preprint arXiv:2502.02339}, 2025.

\bibitem{wu2024mind}
Wenshan Wu, Shaoguang Mao, Yadong Zhang, Yan Xia, Li~Dong, Lei Cui, and Furu
  Wei.
\newblock Mind's eye of llms: Visualization-of-thought elicits spatial
  reasoning in large language models.
\newblock In {\em The Thirty-eighth Annual Conference on Neural Information
  Processing Systems}, 2024.

\bibitem{xai2025grok3}
xAI.
\newblock Grok 3: Advancing real-time reasoning in ai.
\newblock Tech Rep.~1, xAI, 2025.

\bibitem{xia2025geox}
Renqiu Xia, Mingsheng Li, Hancheng Ye, Wenjie Wu, Hongbin Zhou, Jiakang Yuan,
  Tianshuo Peng, Xinyu Cai, Xiangchao Yan, Bin Wang, Conghui He, Botian Shi,
  Tao Chen, Junchi Yan, and Bo~Zhang.
\newblock Geox: Geometric problem solving through unified formalized
  vision-language pre-training.
\newblock In {\em The Thirteenth International Conference on Learning
  Representations}, 2025.

\bibitem{xia2024vision}
Shiyu Xia, Junyu Xiong, Haoyu Dong, Jianbo Zhao, Yuzhang Tian, Mengyu Zhou,
  Yeye He, Shi Han, and Dongmei Zhang.
\newblock Vision language models for spreadsheet understanding: Challenges and
  opportunities.
\newblock {\em arXiv preprint arXiv:2405.16234}, 2024.

\bibitem{xu2025visulogic}
Weiye Xu, Jiahao Wang, Weiyun Wang, Zhe Chen, Wengang Zhou, Aijun Yang, Lewei
  Lu, Houqiang Li, Xiaohua Wang, Xizhou Zhu, et~al.
\newblock Visulogic: A benchmark for evaluating visual reasoning in multi-modal
  large language models.
\newblock {\em arXiv preprint arXiv:2504.15279}, 2025.

\bibitem{yang2024qwen2}
An~Yang, Baosong Yang, Beichen Zhang, Binyuan Hui, Bo~Zheng, Bowen Yu,
  Chengyuan Li, Dayiheng Liu, Fei Huang, Haoran Wei, et~al.
\newblock Qwen2. 5 technical report.
\newblock {\em arXiv preprint arXiv:2412.15115}, 2024.

\bibitem{yang2024thinking}
Jihan Yang, Shusheng Yang, Anjali~W Gupta, Rilyn Han, Li~Fei-Fei, and Saining
  Xie.
\newblock Thinking in space: How multimodal large language models see,
  remember, and recall spaces.
\newblock {\em arXiv preprint arXiv:2412.14171}, 2024.

\bibitem{yao2024mulberry}
Huanjin Yao, Jiaxing Huang, Wenhao Wu, Jingyi Zhang, Yibo Wang, Shunyu Liu,
  Yingjie Wang, Yuxin Song, Haocheng Feng, Li~Shen, et~al.
\newblock Mulberry: Empowering mllm with o1-like reasoning and reflection via
  collective monte carlo tree search.
\newblock {\em arXiv preprint arXiv:2412.18319}, 2024.

\bibitem{yao2023tree}
Shunyu Yao, Dian Yu, Jeffrey Zhao, Izhak Shafran, Thomas~L. Griffiths, Yuan
  Cao, and Karthik~R Narasimhan.
\newblock Tree of thoughts: Deliberate problem solving with large language
  models.
\newblock In {\em Thirty-seventh Conference on Neural Information Processing
  Systems}, 2023.

\bibitem{yao2023react}
Shunyu Yao, Jeffrey Zhao, Dian Yu, Nan Du, Izhak Shafran, Karthik Narasimhan,
  and Yuan Cao.
\newblock React: Synergizing reasoning and acting in language models.
\newblock In {\em International Conference on Learning Representations (ICLR)},
  2023.

\bibitem{ye2025limo}
Yixin Ye, Zhen Huang, Yang Xiao, Ethan Chern, Shijie Xia, and Pengfei Liu.
\newblock Limo: Less is more for reasoning.
\newblock {\em arXiv preprint arXiv:2502.03387}, 2025.

\bibitem{zhan2025open3dvqa}
Weichen Zhan, Zile Zhou, Zhiheng Zheng, Chen Gao, Jinqiang Cui, Yong Li, Xinlei
  Chen, and Xiao-Ping Zhang.
\newblock Open3dvqa: A benchmark for comprehensive spatial reasoning with
  multimodal large language model in open space.
\newblock {\em arXiv preprint arXiv:2503.11094}, 2025.

\bibitem{zhang2024improve}
Ruohong Zhang, Bowen Zhang, Yanghao Li, Haotian Zhang, Zhiqing Sun, Zhe Gan,
  Yinfei Yang, Ruoming Pang, and Yiming Yang.
\newblock Improve vision language model chain-of-thought reasoning.
\newblock {\em arXiv preprint arXiv:2410.16198}, 2024.

\end{thebibliography}

\appendix
\newpage
\section{Appendix Outline}
In the appendix, we provide the following:
\begin{itemize}
    \item Limitations of this study (Appendix~\ref{app: limitations});
    \item Technical details on the construction of Inf-Bench, including our method for encoding graphics, training details, and the implementation specifics of the reasoning enhancement methods we employ (Appendix~\ref{app: Technical Details});
    \item The evaluation setup and detailed error analysis for the Inf-Bench experiments (Appendix~\ref{app: Evaluation Details});
    \item A comprehensive introduction to the graphics, colors, and operations (Appendix~\ref{app: more details of shapes}).
\end{itemize}

\section{Limitations}
\label{app: limitations}
Although we explore various spatial transformation tasks from 2D to 3D based on Shapez and Rubik's Cube, real-world spatial deformations often involve more complex scenarios including continuous transformations, non-rigid objects, and topological changes. Future research should extend to a broader range of deformation types.
Our supervised fine-tuning approach, while yielding improvements on certain tasks, fails to address fundamental limitations in spatial reasoning—particularly the lack of mechanisms for maintaining global consistency and correcting accumulated errors in long sequence dependencies. This suggests the need for innovative architectural designs incorporating physics-inspired attention mechanisms or symbolic reasoning tools.
Additionally, our evaluation primarily focused on model performance at fixed step numbers rather than exploring transfer capabilities between tasks of varying difficulty. Future work should systematically investigate knowledge transfer and generalization abilities from low-dimensional to high-dimensional tasks.
While our experiments demonstrate fundamental limitations in current models' spatial reasoning capabilities, we have not exhaustively explored all possible enhancement techniques. Combining multimodal pretraining with neural-symbolic methods, embodied learning, or external memory mechanisms might provide breakthroughs in this domain.
\section{Technical Details}
\label{app: Technical Details}
\subsection{Detailed Overview of Encoding Methods}
\label{app: encode_method}
This section provides a detailed explanation of the encoding methods mentioned in section~\ref{main:5-3} of the main text.

\subsubsection{2D Encoding Method}
In the 2D task, shapes consist of two components: color and shape type. The color of each shape is represented by a lowercase letter, including red (r), green (g), blue (b), yellow (y), purple (p), cyan (c), colorless (u), and white (w). The shape type is represented by an uppercase letter, including circle (C), rectangle (R), windmill (W), sector (F), and star (S). Each shape comprises four quadrants, arranged in the order of quadrant 1, quadrant 2, quadrant 3, and quadrant 4, starting from the top right quadrant and proceeding clockwise. Each quadrant is represented by a pair of letters: the first letter represents the shape type, and the second letter represents the color. If a quadrant is empty, it is denoted by ``--''. A shape can consist of up to four layers, each comprising four quadrants. The shape encoding starts from the top layer and is arranged in ascending order from bottom to top, with layers separated by a colon. For example, ``Su--Ry--'' represents that in the first layer, quadrant 1 is a colorless star, quadrants 2 and 4 are empty, and quadrant 3 is a yellow rectangle. In this task, the player needs to transform the original shape into the target shape through a series of operations. Shape operations include cutting (removing quadrants 1 and 2, which is equivalent to cutting off the right half of the shape), clockwise rotation by 90\degree  (rotating the quadrants clockwise), counterclockwise rotation by 90\degree  (rotating the quadrants counterclockwise), filling (filling empty quadrants with specified shapes), mirroring (horizontally mirroring the entire shape), and coloring (changing the color of all shapes in the selected layer). These operations gradually transform the shape to match the target.

\subsubsection{2.5D Encoding Method}
In the 2.5D task, shapes not only have multiple layers, each consisting of four quadrants, but the maximum number of layers is four. The shape of each layer is represented by the same rules and arranged in ascending order from bottom to top. Each layer consists of four quadrants, with the order of quadrants being quadrant 1, quadrant 2, quadrant 3, and quadrant 4. Unlike the 2D task, the shapes in 2.5D tasks stack multiple layers on top of each other, with each layer separated by a colon (:) and arranged from the bottom layer to the top layer. The encoding rules for the shape are similar to those in the 2D task, where each layer is composed of four quadrants, and each quadrant is represented by a character pair for shape and color. If a quadrant is empty, it is represented by ``--''. Layers are separated by colons. For example, \{``Layer 1'': ``Su--Ry--'', ``Layer 2'': ``------Wp--''\} indicates that in the first layer, quadrant 1 is a colorless star, quadrant 3 is a yellow rectangle, and in the second layer, quadrant 3 is a windmill. In 2.5D tasks, in addition to basic operations (such as rotation, cutting, filling, coloring), players must also follow physical constraints: for instance, if a quadrant in one layer contains a shape, the corresponding quadrant in the lower layer must also contain a shape to satisfy the physical constraints. Operations include cutting (removing quadrants 1 and 2 from each layer), clockwise rotation by 90\degree  (rotating each layer by 90\degree  clockwise), counterclockwise rotation by 90\degree  (rotating each layer by 90\degree  counterclockwise), stacking (stacking one shape layer on top of another), and coloring (changing the color of all shapes in a layer). These operations allow players to complete shape transformations in a multi-layered structure while adhering to specific physical constraints and operational rules.

\subsubsection{3D Encoding Method}
In the 3D task, shapes are represented as a Rubik's Cube. The Rubik's Cube consists of six faces, each represented by a 3x3 matrix of colors. Each face is composed of 3 rows and 3 columns of colored squares. The faces of the cube are represented by letters: Up (U), Down (D), Left (L), Right (R), Front (F), and Back (B). Each face's color is represented by color characters: yellow (y), white (w), red (r), orange (o), green (g), and blue (b). For example, 
\[
\text{U}: \left[\begin{array}{ccc}
\text{y} & \text{y} & \text{y} \\
\text{y} & \text{y} & \text{y} \\
\text{y} & \text{y} & \text{y}
\end{array}\right]
\]
Indicates that all squares on the upper face are yellow. The operations on the Rubik's Cube include a clockwise rotation of 90\degree  (rotating a face 90\degree  clockwise, changing the colors of that face and adjacent faces), and a counterclockwise rotation of 90\degree  (rotating a face 90\degree  counterclockwise, changing the colors of that face and adjacent faces). The shape encoding method is straightforward, where each face is represented by a 3x3 matrix, and each matrix element represents the color of the square at that position. Players need to transform the Rubik's Cube from its initial state to the target state by performing a series of rotations. The position of each face and the relative position of each square are crucial, requiring precise control of the rotations to ensure the final state matches the target. The rotation operations involve rotating each face by 90\degree , with interdependencies between faces, making the task more complex.

\subsection{Training Details}
\label{app: train_detail}
In our experiment, we perform supervised fine-tuning of the Qwen2.5-VL-72B-Instruct model using the LLaMA-Factory framework to enhance its multimodal processing capabilities further. We chose the LoRA (Low-Rank Adaptation) method as the core optimization strategy. This method efficiently adjusts model parameters through low-rank matrix decomposition, avoiding the high computational cost of full fine-tuning and significantly optimizing the performance of Qwen2.5-VL-72B-Instruct on multimodal tasks (e.g., visual and linguistic integration). Specifically, we set the rank of LoRA to 8, an empirical compromise balancing model expressiveness with reduced parameter updates and lower memory requirements.

To ensure the efficiency and stability of distributed training, we integrate the DeepSpeed ZeRO-3 optimization strategy. This strategy optimizes model parameters, gradients, and optimizer states through partitioning, enabling efficient memory management and computational resource allocation. Thus, it significantly improves training speed and parallelism in multi-GPU or multi-node environments.

For the data, we use two datasets for supervised fine-tuning: shapez and cube. These datasets are designed to evaluate and enhance the model's multimodal understanding capabilities. The shapez dataset is further subdivided into shapez and shapez\_2d subsets. The shapez subset focuses on the stitching, segmentation, and combining 2D images, involving complex reasoning tasks such as spatial relationship analysis and geometric transformations. In contrast, shapez\_2d emphasizes specific variants of 2D images, such as simplified projections or graphic manipulations. The cube dataset, on the other hand, addresses 3D image understanding tasks. By fine-tuning on these diverse datasets, we aim to strengthen the model's generalization ability for visual-linguistic tasks across different dimensions and complexities.

For the training configuration, we set the per-device training batch size to 4, which helps avoid out-of-memory (OOM) errors while maintaining reasonable sample diversity given the limited GPU memory. We further set the gradient accumulation steps to 4, allowing us to simulate larger effective batch sizes despite smaller actual batches, thereby improving gradient estimation accuracy without increasing per-step computation load. To optimize the convergence process, we set the learning rate to 1e-4 (i.e., 0.0001) to maintain stable training. The number of training epochs is set to 3 to ensure efficient iteration under resource constraints.

\subsection{Data Engine}
\label{app: data_engine}
\subsubsection{Shapez Data Engine}
The Shapez Data Engine is primarily used to generate and manage shape structures, supporting the generation of various shapes under specified conditions and transforming these shapes through a series of operations. Its core attributes include two types of fundamental data: shape types (where 'C', 'R', 'W', 'S' represent circle, rectangle, windmill, and star, respectively) and colors ('r', 'g', 'b', 'y', 'p', 'c', 'u', 'w' represent red, green, blue, yellow, purple, cyan, colorless, and white, respectively). The key methods of the engine include \texttt{generate\_shape\_structures} and \texttt{execute\_actions}. The \texttt{generate\_shape\_structures} method is used to generate a set of shape structures, where the number of layers, shape count, color, and consistency of shapes across layers can be specified. During the generation process, shapes and colors are randomly selected, and shapes for each layer are created based on the specified number of layers. Each layer's shape is ensured to contain at least one shape via the \texttt{generate\_non\_empty\_layer} method, and the matching between layers is validated using the \texttt{check\_layer\_validity} method. Finally, the \texttt{generate\_shape\_structure} method combines these layers into a complete shape structure and returns it. The \texttt{execute\_actions} method is used to apply a series of operations (such as rotation, cutting, stacking, coloring, etc.) to the generated shapes. These operations are applied individually through the \texttt{execute} method, and the transformed final shape is returned. The specific algorithm is found in Algorithm~\ref{alg:sde}

\begin{algorithm}
\label{alg:sde}
\normalsize
    \caption{Shapez Data Engine}
    \label{algorithm_shapez_data_engine}
    \begin{algorithmic}[1]
        \State \textbf{Input:} $num\_structures$, $num\_layers$, $color$, $num\_shapes$, $num\_colors$, $all\_the\_same$, $seed$ (optional)
        \State \textbf{Output:} Generated shape structures
        
        \State \emph{\textbf{Phase 0}: Initialize Parameters}
            \If{$seed$ is provided}
                \State Set random seed
            \EndIf
            \State Randomly select $num\_shapes$ from $shapes$
            \State Randomly select $num\_colors$ from $colors$
        
        \State \emph{\textbf{Phase 1}: Generate Shapes and Layers}
            \For{each layer $i$ from $1$ to $num\_layers$}
                \State Generate shape using $color$ and $shape$
                \State Generate a non-empty layer with at least one shape
                \State Check the validity of the layer with respect to the previous layer
            \EndFor

        \State \emph{\textbf{Phase 2}: Create Shape Structure}
            \State Combine valid layers to form a complete shape structure
            \State Repeat the above steps for $num\_structures$ shapes
        
        \State \Return Generated shape structures
    \end{algorithmic}
\end{algorithm}

\subsubsection{Cube Data Engine}
The Cube Data Engine focuses on generating and manipulating 3D cubes. Its design is intended to simulate the polyhedral structure and rotational operations of a Rubik's Cube. Each face of the cube is represented by a 3x3 matrix, with each position's color represented by specific color characters (e.g., yellow 'y', white 'w', red 'r', green 'g', etc.). This engine supports generating different cube configurations based on an initial cube state and executing various rotational operations on the cube. The \texttt{generate\_cube\_structure} method generates an initial cube state by specifying different colors and layouts. The position and color of each face must strictly match to ensure the validity of the cube's structure. The \texttt{execute\_cube\_actions} method performs a series of rotation actions, changing the color layout of the cube's faces. Each rotation operation is implemented through the \texttt{rotate\_face} method, which can rotate the face clockwise or counterclockwise. Each operation affects multiple faces of the cube and their adjacent faces, resulting in a new state. Players can adjust the cube's state according to the task requirements through these operations, ultimately reaching the target configuration. The specific algorithm is found in Algorithm~\ref{alg:cde}

\begin{algorithm}
\label{alg:cde}
\normalsize
    \caption{Cube Data Engine}
    \label{algorithm_cube_data_engine}
    \begin{algorithmic}[1]
        \State \textbf{Input:} $initial\_cube\_state$, $actions$
        \State \textbf{Output:} Modified cube state
        
        \State \emph{\textbf{Phase 0}: Initialize Cube}
            \State Generate initial cube state from $initial\_cube\_state$
        
        \State \emph{\textbf{Phase 1}: Execute Cube Actions}
            \For{each action in $actions$}
                \If{action is a rotation}
                    \State Rotate the specified face of the cube
                \ElsIf{action is another type}
                    \State Execute other cube operations (e.g., color change, swap)
                \EndIf
            \EndFor
        
        \State \Return the Final modified cube state after applying all actions
    \end{algorithmic}
\end{algorithm}

\subsection{Enhancement Details}
\label{app: enhancement_detail}
As mentioned in section~\ref{main:enhance_method} of the main text, we explore several mainstream reasoning enhancement methods, including Chain-of-Thought (COT), Few-shot Learning, Self-reflection, Tool Invocation, and Reasoning and Acting (ReAct). This section provides a detailed explanation of the specific implementation of these methods.

\textbf{COT.} We added the phrase ``Let's think step by step'' to the prompt.

\textbf{Few-shot Learning.} We include an example of a question with the same level of difficulty after the prompt.

\textbf{Self-reflection.} A cycle mechanism of initialization, validation, and correction is introduced to simulate humans' self-reflection process when solving problems. This framework first generates an initial answer, then interacts with external tools to validate the answer's quality, generates self-criticism, and finally refines the answer based on this criticism. Given the model \(M\) and input \(x\), the initial answer is generated by the prompt \(\mathcal{P}\): 
\[
\hat{y}_0 \sim P_M(\cdot \mid \mathcal{P} \oplus x).
\]
Subsequently, the model interacts with external tools to evaluate \(\hat{y}_i\) and generate criticism \(c_i \sim P_M(\cdot \mid \mathcal{P} \oplus x \oplus \hat{y}_i, T)\). These task-specific criticisms can be used to assess various attributes of the output, such as truthfulness, feasibility, or safety. Finally, the model generates an improved answer based on input \(x\), previous output \(\hat{y}_i\), and criticism \(c_i\):
\[
\hat{y}_{i+1} \sim P_M(\cdot \mid \mathcal{P} \oplus x \oplus \hat{y}_i \oplus c_i).
\]
Criticism plays a key role in correcting errors by identifying errors, providing feasible suggestions, or offering reliable justifications through interactions with external tools, guiding the new generation to avoid similar mistakes. This ``validate-correct-validate'' cycle can repeat multiple times until a specific stopping condition is met, such as the validation process satisfying a requirement, reaching the maximum number of iterations, or receiving environmental feedback. The specific algorithm is found in Algorithm~\ref{alg:SELF-REFLECTION}

\begin{algorithm}
\label{alg:SELF-REFLECTION}
\caption{SELF-REFLECTION}
\begin{algorithmic}[1]
\Require Input $x$, prompt $\wp$, model $M$, external tools $T = \{T_1, T_2, \ldots, T_k\}$, number of iterations $n$
\Ensure Corrected output $\hat{y}$ from $M$
\State Generate initial output $\hat{y}_0 \sim P_M(\cdot|\wp \oplus x)$ \Comment{Initialization}
\For{$i \leftarrow 0$ to $n-1$}
    \State Verify $\hat{y}_i$ through interaction with $T$ to obtain critiques $c_i \sim P_M(\cdot|\wp \oplus x \oplus \hat{y}_i, T)$ \Comment{Verification}
    \If{$c_i$ indicates that $\hat{y}_i$ is correct} \Comment{Stopping Criteria}
        \State \Return $\hat{y}_i$
    \EndIf
    \State $\hat{y}_{i+1} \sim P_M(\cdot|\wp \oplus x \oplus \hat{y}_i \oplus c_i)$ \Comment{Correction}
\EndFor
\State \Return $\hat{y}_n$
\end{algorithmic}
\end{algorithm}

\textbf{Tool Invocation.} This is the most straightforward tool usage paradigm, taking the form of an alternating dialogue between the language model and the tools. In this framework, the model generates outputs that include tool invocation requests. The system extracts and executes these calls and then provides the results of these executions back to the model, creating a dialogue loop. The specific algorithm is found in Algorithm~\ref{alg:tool}

\begin{algorithm}
\label{alg:tool}
\caption{TOOL}
\begin{algorithmic}[1]
\Require Input $x$, prompt $\wp$, model $M$, external tools $T = \{T_1, T_2, \ldots, T_k\}$, max steps $n$
\Ensure Final output $y$ from $M$
\State Initialize conversation history $h \leftarrow \wp \oplus x$
\For{$i \leftarrow 1$ to $n$}
    \State Generate model response $r_i \sim P_M(\cdot|h)$
    \If{no tool call detected in $r_i$}
        \State \Return $r_i$
    \EndIf
    \State Extract tool call $(tool\_name, tool\_input)$ from $r_i$
    \State Execute tool: $tool\_output \leftarrow T[tool\_name](tool\_input)$
    \State Update history: $h \leftarrow h \oplus r_i \oplus tool\_output$
\EndFor
\State Generate final response $y \sim P_M(\cdot|h)$
\State \Return $y$
\end{algorithmic}
\end{algorithm}

\textbf{ReAct (Reasoning and Acting).} Here, the model implements a cycle of thinking, acting, and observing, significantly enhancing the problem-solving ability of the language model. It introduces two additional spaces: an action space \(A\), which contains a variety of executable operations, and a thinking space \(L\), for internal reasoning within the model. Unlike direct tool invocation, ReAct emphasizes explicit reasoning steps, enabling the model to ``think'' about the following action. In the ReAct framework, each iteration includes three key steps: 
\begin{enumerate}
    \item First, the model generates a reasoning process in the thinking space, outlining the problem-solving approach;
    \item Second, based on this reasoning, the model decides on the appropriate action to take;
    \item Third, it observes the result of the action and integrates this information into the context.
\end{enumerate}
This structured reasoning-action-observation cycle allows the model to handle complex problems more systematically, especially tasks that require multi-step reasoning and tool collaboration. The specific algorithm is found in Algorithm~\ref{alg:react}

\begin{algorithm}
\label{alg:react}
\caption{REACT}
\begin{algorithmic}[1]
\Require Input $x$, prompt $\wp$, model $M$, external tools $T = \{T_1, T_2, \ldots, T_k\}$, max steps $n$, action space $A$, language space $L$
\Ensure Final output $y$ from $M$
\State Initialize context $c_1 \leftarrow \wp \oplus x$
\For{$i \leftarrow 1$ to $n$}
    \State Generate thought $t_i \sim P_M(\cdot|c_i)$ where $t_i \in L$ \Comment{Thinking}
    \State Update context: $c_i \leftarrow c_i \oplus t_i$
    \State Generate action $a_i \sim P_M(\cdot|c_i)$ where $a_i \in A$ \Comment{Acting}
    \If{$a_i$ is a final answer}
        \State \Return $a_i$
    \EndIf
    \State Execute action to get observation: $o_i \leftarrow Execute(a_i, T)$ \Comment{Observing}
    \State Update context: $c_{i+1} \leftarrow c_i \oplus a_i \oplus o_i$
\EndFor
\State Generate final answer $y \sim P_M(\cdot|c_{n+1})$
\State \Return $y$
\end{algorithmic}
\end{algorithm}
\section{Evaluation Details}
\label{app: Evaluation Details}
\subsection{General Evaluation Setup}
\label{app: General Evaluation Setup}
\begin{figure}[h!]
    \centering
    \vspace{-0.2cm}
      \includegraphics[width=0.3\linewidth]{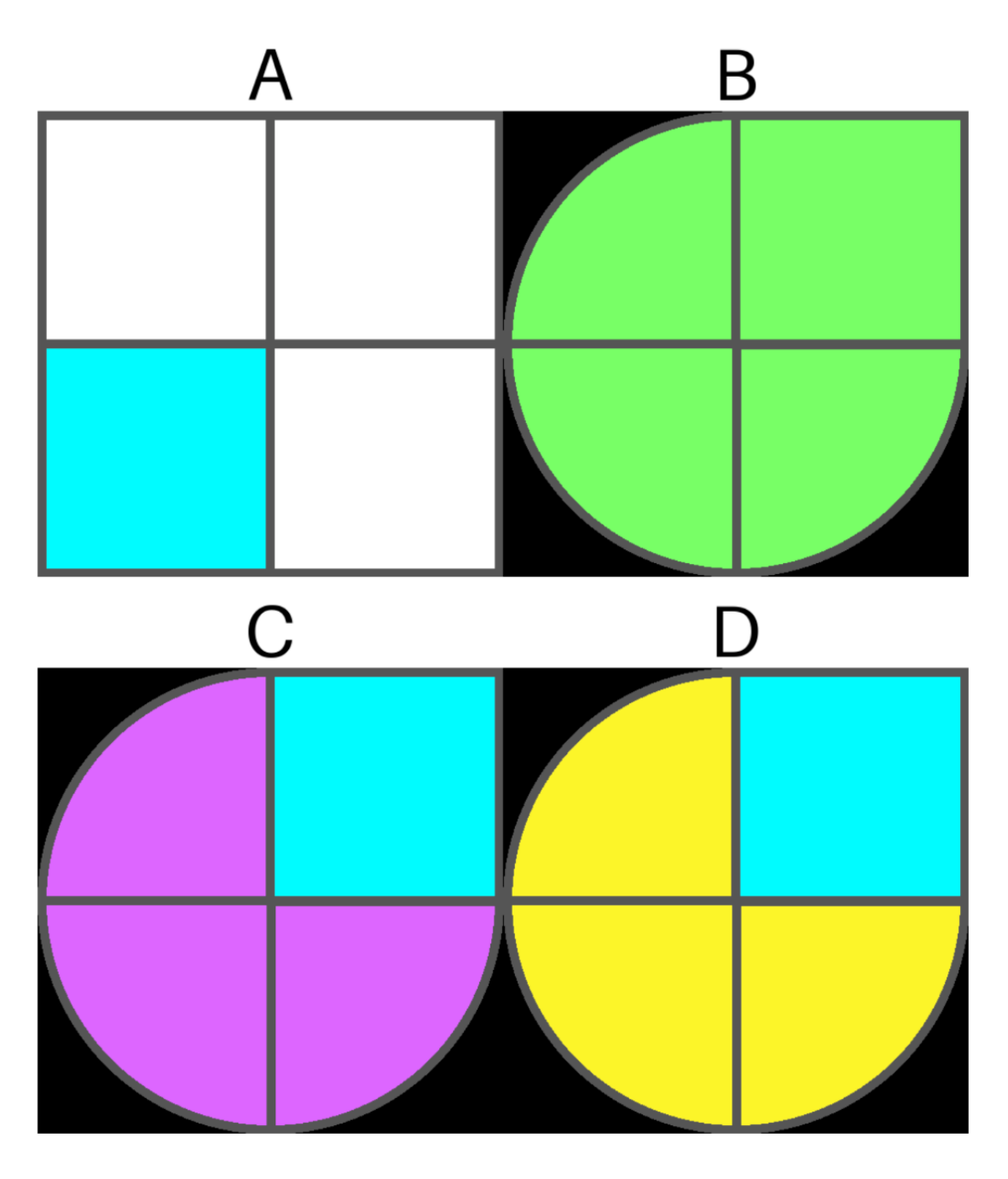}
    \vspace{-0.2cm}
    \caption{
    Example of the Option of Forward Task
    }
    \label{fig: option_example} 
\end{figure}

\begin{figure}[h!]
    \centering
    \vspace{-0.2cm}
      \includegraphics[width=1\linewidth]{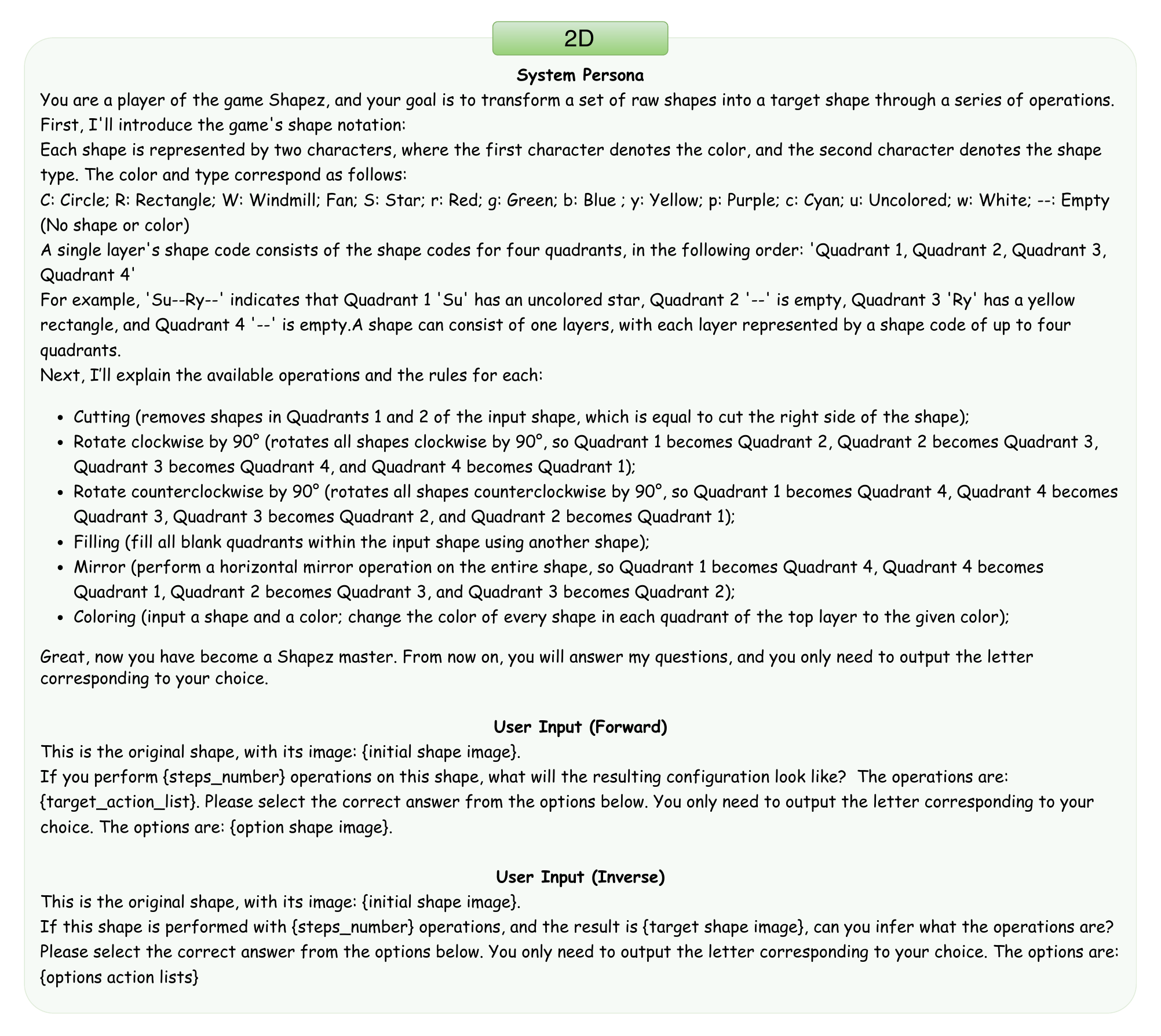}
    \vspace{-0.2cm}
    \caption{
    Prompt of Inf-Bench (2D)
    }
    \label{fig: prompt2d} 
\end{figure}

\begin{figure}[h!]
    \centering
    \vspace{-0.2cm}
      \includegraphics[width=1\linewidth]{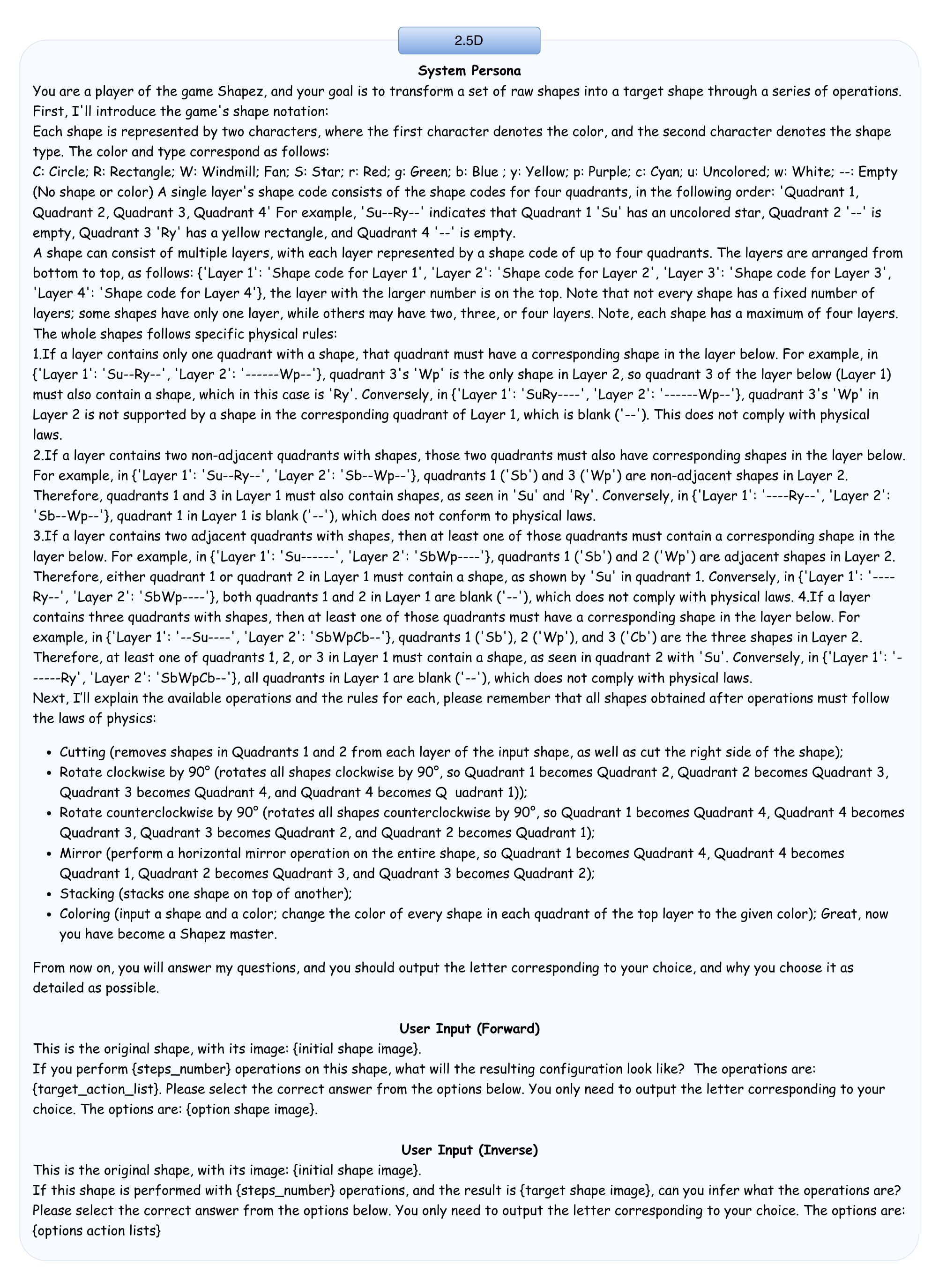}
    \vspace{-0.2cm}
    \caption{
    Prompt of Inf-Bench (2.5D)
    }
    \label{fig: prompt2.5d} 
\end{figure}

\begin{figure}[h!]
    \centering
    \vspace{-0.2cm}
      \includegraphics[width=1\linewidth]{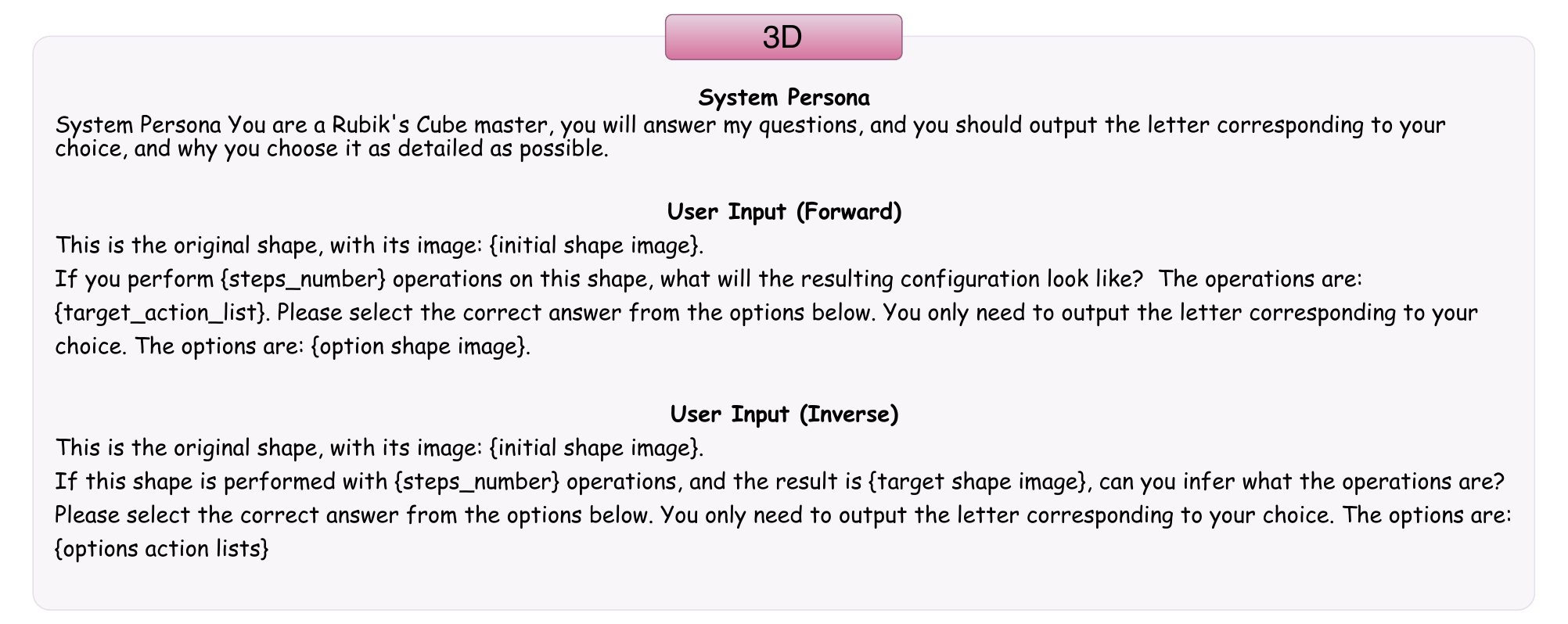}
    \vspace{-0.2cm}
    \caption{
    Prompt of Inf-Bench (3D)
    }
    \label{fig: prompt3d} 
\end{figure}

To ensure reproducibility, unless otherwise specified, we apply a greedy decoding strategy (with a temperature setting of 0, and both top-p and top-k set to 1) for all models. In forward tasks, we present all options within a single image, as shown in Figure~\ref{fig: option_example}. All the prompt used in Inf-Bench is shown in Figures~\ref{fig: prompt2d}, \ref{fig: prompt2.5d}, and \ref{fig: prompt3d}.

\subsection{Human Evaluation Setup}
When evaluating human-level performance on Inf-Bench, human evaluators are allowed unlimited time to answer the questions but can only submit one answer. They receive both the question and the corresponding image simultaneously, and we do not impose any restrictions on their ability to draft responses.

\subsection{Error Analysis Detail and Example}
\label{app: error_ana}

\begin{figure}[h!]
    \centering
    \vspace{-0.2cm}
      \includegraphics[width=1\linewidth]{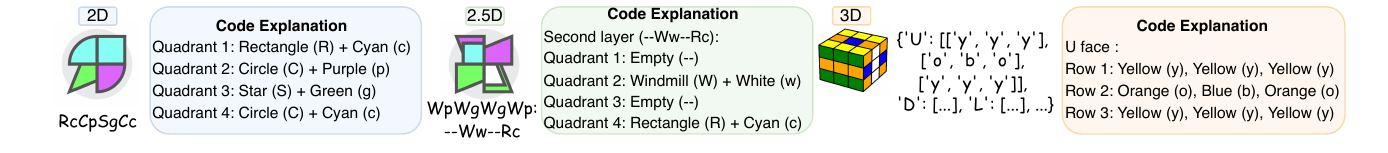}
    \vspace{-0.2cm}
    \caption{
    Examples of Errors
    }
    \label{fig: error_example} 
\end{figure}

For each erroneous case, we classify its primary error into one of four main categories: Pairing Errors, Encoding Errors, Abstracting Errors, and Execution Errors. Suppose an incorrect prediction is attributed to multiple causes. In that case, it is proportionally assigned to each category based on the number of applicable error categories, with \(n\) representing the number of error categories. Examples of these four types of errors are presented in Figure~\ref{fig: error_example}.

\section{More Details about Inf-Bench.}
\label{app: more details of shapes}
\subsection{Shapes and Colors}
\begin{figure}[h]
    \centering
    \vspace{-0.2cm}
      \includegraphics[width=0.95\linewidth]{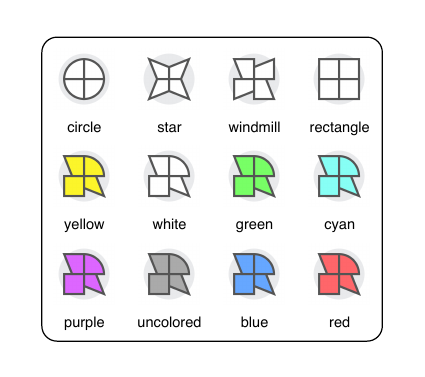}
    \vspace{-0.2cm}
    \caption{
    All Shapes and Colors
    }
    \label{fig: all_shapes} 
\end{figure}
In the Shapez game, shapes can be combined using different types and colors. The shape types include circles, rectangles, windmills, sectors, and stars. Each shape can be represented in various colors, such as red, green, blue, yellow, purple, cyan, colorless, and white. All of these are illustrated in Figure~\ref{fig: all_shapes}.
\subsection{Operations}
\textbf{All Operations in 2D and 2.5D}  

\begin{itemize}
    \item \textbf{Cutting}: Removing a portion of the shape, typically by cutting off the right-side quadrants (quadrant 1 and quadrant 2). For example, cutting a shape with four quadrants into two parts from the right side.
    \item \textbf{Clockwise Rotation 90\degree }: Rotating all quadrants of the shape 90\degree  clockwise, with quadrant 1 becoming quadrant 2, quadrant 2 becoming quadrant 3, quadrant 3 becoming quadrant 4, and quadrant 4 becoming quadrant 1.
    \item \textbf{Counterclockwise Rotation 90\degree }: Rotating all quadrants of the shape 90\degree  counterclockwise, with quadrant 1 becoming quadrant 4, quadrant 4 becoming quadrant 3, quadrant 3 becoming quadrant 2, and quadrant 2 becoming quadrant 1.
    \item \textbf{Filling}: Filling empty quadrants with a specified shape, such as filling the blank quadrant with a rectangle or circle.
    \item \textbf{Mirror}: Performing a horizontal mirror operation on the entire shape, swapping quadrants 1 and 4, and quadrants 2 and 3. The mirror operation reverses the left and right sides of the shape.
    \item \textbf{Coloring}: Changing the color of all quadrants in the shape allows a new color to be applied to the shapes in a specified layer or across all layers.
    \item \textbf{Stacking}: This operation is primarily used in the 2.5D task, where one shape layer is stacked on top of another, forming a multi-layered structure.
\end{itemize}

\textbf{All Rotation Operations in 3D}  

\textit{Face Rotations  }

These operations rotate one of the six faces of the Rubik's Cube. Each face can be rotated clockwise, counterclockwise, or by 180\degree.

\begin{itemize}
    \item \textbf{R}: Rotate the right face 90\degree  clockwise
    \item \textbf{U}: Rotate the upper face 90\degree  clockwise
    \item \textbf{F}: Rotate the front face 90\degree  clockwise
    \item \textbf{D}: Rotate the down face 90\degree  clockwise
    \item \textbf{L}: Rotate the left face 90\degree  clockwise
    \item \textbf{B}: Rotate the back face 90\degree  clockwise
    \item \textbf{R'}: Rotate the right face 90\degree  counterclockwise
    \item \textbf{U'}: Rotate the upper face 90\degree  counterclockwise
    \item \textbf{F'}: Rotate the front face 90\degree  counterclockwise
    \item \textbf{D'}: Rotate the down face 90\degree  counterclockwise
    \item \textbf{L'}: Rotate the left face 90\degree  counterclockwise
    \item \textbf{B'}: Rotate the back face 90\degree  counterclockwise
    \item \textbf{R2}: Rotate the right face 180\degree 
    \item \textbf{U2}: Rotate the upper face 180\degree 
    \item \textbf{F2}: Rotate the front face 180\degree 
    \item \textbf{D2}: Rotate the down face 180\degree 
    \item \textbf{L2}: Rotate the left face 180\degree 
    \item \textbf{B2}: Rotate the back face 180\degree 
\end{itemize}

\textit{Wide Layer Rotations}  

These operations rotate two layers of the Rubik's Cube at once, affecting adjacent layers.

\begin{itemize}
    \item \textbf{r}: Rotate the right two layers 90\degree  clockwise
    \item \textbf{u}: Rotate the upper two layers 90\degree  clockwise
    \item \textbf{f}: Rotate the front two layers 90\degree  clockwise
    \item \textbf{d}: Rotate the down two layers 90\degree  clockwise
    \item \textbf{l}: Rotate the left two layers 90\degree  clockwise
    \item \textbf{b}: Rotate the back two layers 90\degree  clockwise
    \item \textbf{r'}: Rotate the right two layers 90\degree  counterclockwise
    \item \textbf{u'}: Rotate the upper two layers 90\degree  counterclockwise
    \item \textbf{f'}: Rotate the front two layers 90\degree  counterclockwise
    \item \textbf{d'}: Rotate the down two layers 90\degree  counterclockwise
    \item \textbf{l'}: Rotate the left two layers 90\degree  counterclockwise
    \item \textbf{b'}: Rotate the back two layers 90\degree  counterclockwise
    \item \textbf{r2}: Rotate the right two layers 180\degree 
    \item \textbf{u2}: Rotate the upper two layers 180\degree 
    \item \textbf{f2}: Rotate the front two layers 180\degree 
    \item \textbf{d2}: Rotate the down two layers 180\degree 
    \item \textbf{l2}: Rotate the left two layers 180\degree 
    \item \textbf{b2}: Rotate the back two layers 180\degree 
\end{itemize}

\textit{Middle Layer Rotations}  

These operations rotate the middle layers of the Rubik's Cube, affecting rotations between faces.

\begin{itemize}
    \item \textbf{M}: Rotate the middle layer between the left and right faces 90\degree  clockwise
    \item \textbf{S}: Rotate the middle layer between the front and back faces 90\degree  clockwise
    \item \textbf{E}: Rotate the middle layer between the top and bottom faces 90\degree  clockwise
    \item \textbf{M'}: Rotate the middle layer between the left and right faces 90\degree  counterclockwise
    \item \textbf{S'}: Rotate the middle layer between the front and back faces 90\degree  counterclockwise
    \item \textbf{E'}: Rotate the middle layer between the top and bottom faces 90\degree  counterclockwise
    \item \textbf{M2}: Rotate the middle layer between the left and right faces 180\degree 
    \item \textbf{S2}: Rotate the middle layer between the front and back faces 180\degree 
    \item \textbf{E2}: Rotate the middle layer between the top and bottom faces 180\degree 
\end{itemize}

\end{document}